\documentclass[a4paper,10pt]{article}
\usepackage[utf8]{inputenc}
 \usepackage[english]{babel}
\usepackage{graphicx}
\usepackage{a4wide}

\usepackage{amsmath}
\usepackage{amsfonts}
\usepackage{lipsum}
\usepackage{subfig}
\usepackage{pifont}
\usepackage{nicefrac}
\usepackage{multirow}
\usepackage{array}
\usepackage{hhline}
\usepackage{tabularx}
\usepackage{colortbl}
\usepackage{color}
\usepackage{hhline}
\usepackage{natbib}
\usepackage{colortbl}
\usepackage{hhline}
\usepackage{multirow}
\usepackage{longtable}
\usepackage{rotating}
\usepackage{enumerate}
\usepackage{algorithm2e}
\usepackage{endfloat}
\usepackage{tikz}
\usepackage{authblk}
\usepackage[notlof]{tocbibind}
\usepackage{pdflscape}
\usetikzlibrary{decorations.pathreplacing}
\usetikzlibrary{shapes.misc}
\usetikzlibrary{shapes, snakes}
\usepackage{titlesec}
\usepackage{gensymb}

\titleformat{\paragraph}
{\normalfont\normalsize\bfseries}{\theparagraph}{1em}{}
\titlespacing*{\paragraph}
{0pt}{3.25ex plus 1ex minus .2ex}{1.5ex plus .2ex}

\makeatletter
\newcommand*{\rom}[1]{\expandafter\@slowromancap\romannumeral #1@}
\makeatother

\graphicspath{{Figures/}{/media/ricardo/1401-1FFE/Documents/Writing/rt-Paper/Figures/}} 

\title{Measuring the functional connectome ‘on-the-fly’ --- towards
a new control signal for fMRI-based brain-computer interfaces}
\author[1]{Ricardo Pio Monti}
\author[2,3]{Romy Lorenz}
\author[1]{Christoforos Anagnostopoulos}
\author[2]{Robert Leech}
\author[1,4]{Giovanni Montana\footnote{Corresponding author: \tt{giovanni.montana@kcl.ac.uk}}}
\affil[1]{Department of Mathematics, Imperial College London, London SW7 2AZ, UK}
\affil[2]{Computational, Cognitive and Clinical Neuroimaging Laboratory, Imperial College London, The Hammersmith Hospital, London W12 0NN, UK}
\affil[3]{Department of Bioengineering, Imperial College London, London SW7 2AZ, UK}
\affil[4]{Department of Biomedical Engineering, King's College London, St Thomas' Hospital, London SE1 7EH, UK}
\date{} 

\begin{document}

\maketitle
\begin{abstract}

There has been an explosion of interest in functional 
Magnetic Resonance Imaging (MRI) during the past two decades. 
Naturally, this has been accompanied by 
many major advances in the understanding of the human connectome.
These advances have served to pose 
novel challenges as well as open new avenues for research. One of the most promising and exciting
of such avenues is the study of functional MRI in real-time.
Such studies have recently gained momentum and have been applied in a wide variety of 
settings; ranging from 
training of healthy subjects to self-regulate neuronal activity to
being suggested as potential treatments for clinical populations. 
To date, the vast majority of these studies have focused on a single region 
at a time. This is due in part to the many challenges faced when
estimating dynamic functional connectivity networks in real-time.
In this work we propose a novel methodology with which to accurately track changes in functional 
connectivity networks in real-time.
We adapt the recently proposed SINGLE algorithm for estimating sparse
and temporally homogeneous dynamic networks to be applicable in real-time.
The proposed method is applied to motor task
data from the Human Connectome Project as 
well as to real-time data obtained while exploring a virtual environment. We show that the algorithm
is able to estimate significant task-related changes in network structure quickly enough 
to be useful in future 
brain-computer interface 
applications. 

\end{abstract}

\vspace{15mm}
{\footnotesize
\section*{{\small{Acknowledgements}}}
The authors wish to thank Dr. Gregory Scott for providing help with the programming and preparation of the 
virtual reality platform used in the final application.

Data were provided (in part) by the Human Connectome Project, 
WU-Minn Consortium (Principal Investigators: David Van Essen and Kamil Ugurbil; 1U54MH091657) 
funded by the 16 NIH Institutes and Centers that support the NIH Blueprint for Neuroscience Research; 
and by the McDonnell Center for Systems Neuroscience at Washington University
}

\newpage
\section{Introduction}
\label{sec--rtSINGLE}

The notion of mind-controlled technology has been fueled by rapid advances in brain imaging over the last century. 
Since their early beginnings in the 1970s \citep{vidal1973toward, vidal1977real}, 
brain-computer interfaces (BCI) have evolved 
into ``one of the fastest growing areas of scientific research'' \citep{mak2009clinical}.
Formally, a BCI is a system that measures brain activity and uses it
to replace, restore, enhance, supplement, or improve normal output channels of peripheral 
nerves and muscles \citep{wolpaw2011brain}. 
While initial BCI research sought to enable communication with locked-in patients, 
its motivations and aspirations have since grown to encompass 
a wide range of applications such as motor rehabilitation \citep{birbaumer2007brain}
and movement restoration for the paralyzed  \citep{muller2005eeg}. Moreover,
BCI serves not only to bypass the brain's natural motor output, but it can also
transform brain signals into sensory input that can subsequently be used to modify
cognitive state or behavior, a process referred to as neurofeedback \citep{ruiz2014real}.

Neurofeedback is a specific form of biofeedback whereby subjects
are made aware of their brain activity in real-time. 
Such methods have been successfully employed to train the self-regulation of brain activity 
\citep{wolpaw2002brain, christopher2008applications, birbaumer2009neurofeedback, weiskopf2012real}.
Due to its 
inexpensive equipment and high temporal resolution, early neurofeedback research focused mainly
on electroencephalography (EEG) as the preferred recording method. 
Yet, the lack of precise localization and the limited
access to deep cortical and sub-cortical areas have limited future progress of EEG-based research. In contrast,
functional magnetic resonance imaging (fMRI) has relatively high spatial resolution and whole brain coverage. Recent 
technical and methodological advances in acquisition and analysis have made
real-time fMRI (rt-fMRI) a viable alternative when performing 
neurofeedback studies \citep{sitaram2010real}.

A recent literature review by \cite{ruiz2014real} highlighted that the vast majority of studies to-date
have employed 
rt-fMRI for training healthy individuals to volitionally control BOLD activity in specific brain areas. Such region 
of interest (ROI) based neurofeedback training has been reported to dynamically reconfigure functional brain networks
\citep{haller2013dynamic}
and reinforce effective connectivity \citep{ruiz2013acquired, lee2012real}. 
However, while such ROI based studies have provided fundamental insights into functional architecture and 
cognition, they do no take into consideration 
the fact that complex cognitive processes are not limited to single brain regions but rather result from interactions
between brain regions and between networks of regions 
\citep{sporns2004organization, bressler2010large, koush2013connectivity, ruiz2014real}.
The next frontier for rt-fMRI studies
corresponds to providing accurate and timely feedback to subjects based on
entire functional connectivity networks as opposed to single ROIs, thereby
providing a far richer description of a subjects' brain state.

To date, there have been only a limited number of studies involving neurofeedback based on connectivity.
One of the first studies to demonstrate self-regulation of functional connectivity networks was 
performed by \cite{ruiz2014real}. Here a sliding window was used to provide subjects with a visual measure of 
functional connectivity between two ROIs. 
\cite{zilverstand2014windowed} 
performed an offline analysis showing that windowed correlations provide valuable information relating
to task difficulty. Moreover, such measures of functional connectivity were shown to be more informative 
than univariate activation-based approaches. In a hypothesis-led study, \cite{koush2013connectivity}
presented
a near real-time approach in which subjects learned to modulate the effective connectivity 
(assessed using dynamic causal modeling) between two pairs of ROIs. 

These studies suggest that rt-fMRI connectivity is a useful
tool for neurofeedback. However, such an endeavor presents considerable methodological challenges.
Firstly, due to the nature of neurofeedback the 
resulting time series are expected to be non-stationary. The accurate estimation of 
non-stationary functional connectivity networks in an offline setting is a difficult 
problem in its own right and has recently received considerable attention 
\citep{bassett2011dynamic, allen2012tracking, DCR, Monti2014, davison2015brain}.
In this work we look to address this issue by extending recently proposed methods from
the offline domain to the real-time domain. Second, due to potentially rapid changes 
that may occur in a subjects’ functional connectivity the proposed method must be both 
computationally efficient as well as highly adaptive to change. In order to satisfy the
latter, the proposed method must be capable of accurately estimating functional 
connectivity networks using only a reduced (and adequately re-weighted) subset of current and past BOLD measurements. 

To address these challenges, we first propose the use of  
exponentially weighted moving average (EWMA) models as well as more general adaptive forgetting techniques.
This decision
is motivated
by the superior statistical properties of such approaches \citep{lindquist2014evaluating}
as well as the need to ensure that the proposed methods are as adaptive as possible.
We then extend the recently proposed 
Smooth Incremental Graphical Lasso Estimation (SINGLE) algorithm \citep{Monti2014}. 
Here functional relationships between pairs of nodes are estimated 
using partial correlations as opposed to Pearson's correlation.
Partial correlations are employed as they estimates pairwise correlations between nodes once 
the effects of all 
other nodes have been removed and have been shown to be 
better suited to 
detecting changes in network structure \citep{smith2011network, marrelec2009}. 
We are able to re-cast the estimation of a new functional connectivity network as a 
convex optimization problem which
can be quickly and efficiently solved in real-time. 

The remainder of this manuscript is organized as follows:
in Section \ref{sec--rtSINGLEmethods} we introduce and describe the 
proposed method. 
In Section \ref{sec--rtSINGLEsims} we demonstrate the capabilities of the proposed method 
via  a series of simulations. 
Finally, in Sections \ref{sec--HCPapp} and \ref{sec--Minecraftapp} we present two applications of the
proposed algorithm. 
The first corresponds to a 
proof-of-concept study involving task-based data from the Human Connectome Project \citep{elam2014human, van2012human}.
While this data is not implicitly real-time, it may be treated 
as such to demonstrate the capabilities of the proposed method. 
The second application involves real-time fMRI data obtained from a
single individual exploring a virtual environment.

\section{Methods}
\label{sec--rtSINGLEmethods}

We assume we have access to a stream of multivariate fMRI measurements across $p$ nodes where each 
node represents a region of interest (ROI). 
We write $X_t \in \mathbb{R}^{1\times p}$ to denote the BOLD measurements at the $t^{\mbox{\small th}}$
time point
across $p$ ROIs; thus $X_{t,i}$ corresponds to the BOLD measurement of the $i$th 
node at time $t$.
In this work we are interested 
in sequentially using all
observations up to and including $X_t$ to 
recursively learn the underlying functional connectivity 
networks.
At time $t+1$ it is assumed we receive a new observation $X_{t+1}$, which 
we use to update our network estimates accordingly.
Throughout the remainder of this  manuscript it is assumed that 
each $X_t$ follows a multivariate Gaussian distribution, $ X_t \sim \mathcal{N}(\mu_t, \Sigma_t)$, where
both the 
mean and covariance structure are assumed to vary over time. 

The functional connectivity network at time $t$ can be estimated by learning the 
corresponding precision (inverse covariance) matrices, $\Sigma_t^{-1}=\Theta_t$. 
Such approaches have been employed extensively in neuroimaging applications
\citep{varoquaux2010brain, smith2011network, ryali2012estimation}
and have also recently been proposed to estimate 
time-varying estimates of functional connectivity networks \citep{allen2012tracking, DCR, Monti2014}.
Here $\Theta_t$ encodes the 
partial correlations as well as the conditional independence structure at time $t$. 
We then encode $\Theta_t$
as a graph, $G_t$, where the presence of an edge implies a non-zero
entry in the corresponding entry of the precision matrix \citep{lauritzen1996graphical}. 

Therefore, our aim is to estimate an increasing sequence of functional
connectivity networks, $\{G\} = \{G_1, \ldots, G_t, \ldots\}$
where each $G_t$ captures the
functional connectivity structure at the $t$th observation.
We wish for the proposed method to have the following properties:
\begin{enumerate}[(a)]
 \item \textbf{Real-time}: first and foremost,
 networks should be estimated in real-time in order to provide subjects with
feedback in a timely manner. This is of great importance as subjects require prompt feedback in order to
successfully learn self-regulation.
\item \textbf{Adaptivity}: we are particularly interested in the 
changes caused by the direct interaction with subjects while they are in the scanner. As such, it is crucial to 
be able to rapidly quantify changes in functional connectivity structure once these have occurred.
The need for highly adaptive estimation methodologies is further exacerbated
by the lagged nature of the hemodynamic response function, 
where changes in functional measurements typically occur six seconds after performing a task \citep{laconte2007real}.
\item \textbf{Accuracy}: we also wish to accurately estimate network structure
over time. This involves both the accurate estimation of network connectivity at each 
time point as well as the temporal evolution of pairwise relationships over time.
That is to say, estimated networks should provide accurate
representations of the true underling functional connectivity structure at any
point in time as well as accurately describing how networks evolve over time.
\end{enumerate}

Arguably the dominant approach used to obtain adaptive functional connectivity estimates  
involves the use of sliding windows
\citep{hutchison2013dynamic}
and this also holds true
in the rt-fMRI setting \citep{gembris2000functional, esposito2003real, ruiz2014real, zilverstand2014windowed}. 
Such methods are able to obtain adaptive functional connectivity estimates in real-time by only considering 
a fixed number of past observations, defined as the window. 
Using only the observations within the predefined window, a local (i.e., adaptive) estimate
of functional connectivity is obtained. 
A natural extension of sliding windows are exponentially weighted moving average (EWMA) models, 
where observations are downweighted based on their chronological proximity --- thereby giving
more recent observations greater importance \citep{hunter1986exponentially}. 
In such models, information from past observations is discarded at a constant rate determined by a fixed forgetting factor.
Furthermore, adaptive forgetting methods can be seen as a generalization of EWMA models
where the rate at which previous information is discarded is allowed to vary depending on the nature of the data
\citep{haykin2008adaptive}.
This allows such 
algorithms to actively reduce the rate at which past information is discarded while networks remain relatively stable --- resulting in 
more reliable network estimation --- while also 
adapting rapidly to changes by
increasing the rate at which past information is discarded in the presence of changes.
These three methods,  as well as their relationship, are discussed in detail in Section \ref{sec--rtSINGLEEWMA}.

In order to ensure estimated networks provide an accurate representation of true functional connectivity networks we encourage 
two 
properties in estimated functional connectivity networks, $\{G\}$. The first is sparsity; 
while functional connectivity networks are theorized to have evolved to achieve
high efficiency of information transfer at a low connection cost \citep{bullmore},
the main motivation behind the introduction of sparsity here is 
statistical.
Formally, the introduction of sparsity 
ensures the estimation problem remains feasible
when the number of relevant observations falls below the number of parameters to estimate
\citep{michel2011total, ryali2012estimation}. In the presence 
of rapid changes the number of relevant observations falls drastically. In such a scenario,
sparse methods are able to guarantee 
the accurate estimation of functional 
connectivity networks without compromising the adaptivity of 
the proposed method. 
The second property we wish to encourage is temporal homogeneity; 
from a neurofeedback perspective we expect changes in functional connectivity structure to occur predominantly when
paradigm changes occur (e.g., a subject begins performing a different task). Thus we expect network structure to remain constant within a 
neighbourhood of any observation but to vary over a longer period of time. We therefore encourage sparse innovations in network structure over time, 
ensuring that a change in connectivity is only reported when strongly substantiated by evidence in the data. 
Finally, real-time performance is achieved by casting the estimation $G_t$ as a convex optimization problem which can 
be efficiently solved. 

The task of estimating $\Theta_t$ in real time can be broken into two independent steps. First, an updated estimate 
of the sample covariance, $S_t$, is calculated. 
We propose two methods with which an adaptive and accurate estimate of $S_t$ can be obtained: EWMA models 
and adaptive forgetting. 
In a second step,  
the corresponding precision matrix, $\Theta_t$, is estimated given the sample covariance. 
This is achieved by extending the 
recently proposed Smooth Incremental Graphical Lasso Estimation (SINGLE) algorithm \citep{Monti2014} from the
offline domain to the real-time domain. 

The remainder of this section is organized as follows: in Section \ref{sec--rtSINGLEEWMA} we
describe how adaptive estimates of the sample covariance can be obtained in real-time via the use of 
EWMA models or adaptive forgetting techniques. 
In Section \ref{sec--rtSINGLEAlgo}
we outline the optimization algorithm employed. Parameter selection is discussed in Section \ref{sec--rtSINGLEParams}.

\subsection{Real-time, adaptive covariance estimation}
\label{sec--rtSINGLEEWMA}

The estimation of functional connectivity networks is fundamentally a statistical challenge \citep{fristonFC}
which is often studied by quantifying the pairwise  correlations across 
various ROIs. 
Such approaches correspond directly to estimating and studying the covariance structure.
When the functional time series is assumed to be stationary, this coincides with studying the sample covariance matrix 
for the entire dataset. However, in the case of rt-fMRI studies we are faced with data that is
inherently non-stationary. Moreover, we have the additional constraint that 
data arrives sequentially over time, implying that information from new observations must be efficiently incorporated
to update network estimates.

Addressing the non-stationary nature of the data is a challenging problem, even in the offline setting.
Approaches such as change-point detection have been proposed \citep{robinson2010change, DCR}, however the 
most widespread methodology involves the use of sliding windows or generalizations thereof. 
The advantage of such methods is that they
are conceptually simple and can be 
easily extended to 
the real-time scenario as we describe below. A sliding window may be used to obtain a local estimate of the 
sample covariance, $S_t$, at time $t$ as follows: 
\begin{equation}
\label{SWcov}
S_t = \frac{1}{h} \sum_{i=0}^{h-1} (X_{t-i} - \bar x_t)^T (X_{t-i} - \bar x_t),  
\end{equation}
where $\bar x_t$ is the mean of all observations falling within the sliding window and parameter $h$
is the length of the sliding window. It follows that $h$ 
determines the period of time over which previous observations are considered and will directly affect the adaptivity of 
the proposed algorithm.

A natural extension of sliding windows is the use of an exponentially weighted moving averages (EWMA), first introduced by \cite{roberts1959control}.
Here observations are re-weighted according to their chronological proximity. 
The rate at which past information is discarded is determined by a fixed forgetting factor, $r \in (0,1]$.
In this way, EWMA models are able to give greater importance 
to more recent observations thus increasing the adaptivity of the resulting algorithm. Moreover, as described in \cite{lindquist2014evaluating},
these methods enjoy superior statistical properties when compared to sliding window algorithms.
EWMA models thereby provide a conceptually simple and robust method with which to handle a wide range of non-stationary processes. They are also
particularly well suited to the real-time setting as we discuss below.

For a given forgetting factor, $r \in (0,1]$, the estimated mean at time $t$ can be recursively defined as:
\begin{align}
\label{EWMAmu}
 \bar x_t &= \left (1- \frac{1}{\omega_t} \right )  \bar x_{t-1} +  \frac{1}{\omega_t}  X_t
\end{align}
where $\omega_t$ is a normalizing constant which is calculated as:
\begin{align}
\label{EWMAomega}
\omega_t & = \sum_{i=1}^t r^{t-i} = r \cdot \omega_{t-1} + 1. 
\end{align}
The sample covariance at time $t$ is subsequently defined as\footnote{
We note that equations (\ref{EWMApi}) and (\ref{EWMAcov}) are equivalent to estimating the sample covariance in the more intuitive manner 
$S_t = \left (1- \frac{1}{\omega_t} \right )  S_{t-1} +  \frac{1}{\omega_t}  (X_t - \bar x_t) (X_t - \bar x_t)^T $, however we choose
to follow this parameterization in order to simplify future discussion.}:
\begin{align}
\label{EWMApi}
\Pi_t &= \left (1- \frac{1}{\omega_t} \right )  \Pi_{t-1} +  \frac{1}{\omega_t}  X_t X_t^T\\
\label{EWMAcov}
S_t &= \Pi_t - \bar x_t \bar x_t^T
\end{align}

From equations (\ref{EWMAmu}) and (\ref{EWMApi}) we note
that past observations gradually receive less importance. This is a contrast to sliding windows, where all
observations receive equal weighting. 
It follows that the choice of parameter $r$ determines
the rate at which information from previous observations is discarded and is directly related to the 
adaptivity of the proposed method.
This can be seen by considering the extreme cases where and $r=1$. 
Here we have that $\omega_t=t$ and consequently that $\bar x_t$ and $S_t$ correspond 
to the sample mean and covariance estimated in an offline setting (using all observations up to time $t$). As a result equal 
importance is given to all observations, leading to reduced adaptivity to changes. As the value of $r$ is reduced, greater importance is 
given to more recent observations resulting in an increasingly adaptive estimate. Of course, as the value of $r$ decreases the 
estimated mean and covariance become increasing susceptible to outliers and noise. 
The choice of $r$ therefore constitutes a trade-off between
adaptivity and stability. 

Much like the length of the sliding window, $h$, the choice of $r$ essentially determines the effective sample size used to estimate both 
$\bar x_t$ and $S_t$. Therefore the same logic applies when choosing both $r$ and $h$: the value must be sufficiently large so as to allow
robust estimation of the sample covariance without becoming too large \citep{Sakoglu2010}. This is discussed further in Section \ref{sec--rtSINGLEParams}.

We further note that equations (\ref{EWMAomega}) - (\ref{EWMAcov})  make it clear how an real-time implementation of such methods
would work. In practice only the most recent estimates of $\bar x_t$ and $ \Pi_t$ would be stored together with $\omega_t$
from which new updates can efficiently be calculated.

\subsubsection{Adaptive Forgetting}

The use of both a sliding window or an EWMA model requires the specification of a fixed window length, $h$, or forgetting factor, $r$. 
The choice of these parameters makes implicit assumptions relating to the dynamics of the available data. 
It follows that large choices of $r$ and $h$ reflect an assumption that the data is close to being stationary. In such a scenario, 
large choices of $r$ and $h$ allow for accurate estimation of 
sample covariance matrices by adequately leveraging information across a wide range of observations. By the same token,
smaller choices of $h$ and $r$ correspond to an assumption that the statistical properties of the data are changing at a faster rate.

\begin{figure}[h!]
\centering
\begin{tikzpicture}
 \draw [thick, ->] (0, 0) -- (7.5, 0);
 \node at (3.75, -.5) {$t$};
 \draw [thick, ->] (0,0) -- (0, 1.75); \draw [gray] (0.25, 0.25) -- (0.35, 0.35);
 \draw [gray] (0.25, 0.35) -- (0.35, 0.25);
 \node at (-.5, 0.875) {$X_t$};

 \draw [ thick, blue] (0.1, .5) -- (2, .5);
 \draw [thick, blue] (2, 1) -- (7.5, 1);
 \draw [ thick, blue] (2, 1) -- (2, .5);
 
 \draw [gray] (0.25, 0.25) -- (0.35, 0.35);
 \draw [gray] (0.25, 0.35) -- (0.35, 0.25);
 
 \draw [gray] (0.75, 0.7) -- (0.85, 0.8);
 \draw [gray] (0.75, 0.8) -- (0.85, 0.7);

 \draw [gray] (1.25, 0.65) -- (1.35, 0.55);
 \draw [gray] (1.35, 0.65) -- (1.25, 0.55);
 
 \draw [gray] (1.75, 0.425) -- (1.85, 0.325);
 \draw [gray] (1.85, 0.425) -- (1.75, 0.325);
 
 \draw [gray] (2.25, 1.225) -- (2.35, 1.325);
 \draw [gray] (2.35, 1.225) -- (2.25, 1.325);

 \draw [gray] (2.75, 1.025) -- (2.85, 1.125);
 \draw [gray] (2.85, 1.025) -- (2.75, 1.125); 
 
 \draw [gray] (3.35, .9) -- (3.25, 1.);
 \draw [gray] (3.25, .9) -- (3.35, 1.);  
 
 \draw [gray] (3.75, 1.05) -- (3.85, 1.15);
 \draw [gray] (3.85, 1.05) -- (3.75, 1.15);  
 
 \draw [gray] (4.35, .825) -- (4.25, .925);
 \draw [gray] (4.25, .825) -- (4.35, .925);  
 
 \draw [gray] (4.75, .75) -- (4.85, .85);
 \draw [gray] (4.85, .75) -- (4.75, .85);
 
 \draw [gray] (5.35, 1.08) -- (5.25, 1.18);
 \draw [gray] (5.25, 1.08) -- (5.35, 1.18);  

 \draw [gray] (5.75, .95) -- (5.85, 1.05);
 \draw [gray] (5.85, .95) -- (5.75, 1.05); 
 
 \draw [gray] (6.35, 1.06) -- (6.25, 1.16);
 \draw [gray] (6.25, 1.06) -- (6.35, 1.16); 
 
 \draw [gray] (6.75, .985) -- (6.85, 1.085);
 \draw [gray] (6.85, .985) -- (6.75, 1.085); 
 
 \draw [thick, dashed] (1.50, .25) -- (1.5, 1.5);
 \draw [thick, dashed] (2.50, .25) -- (2.5, 1.5);
 \draw [thick,decorate,decoration={brace,amplitude=2pt},xshift=0.4pt,yshift=-0.4pt](1.5,1.55) -- (2.5,1.55) node[black,midway,yshift=-0.6cm] {};
 \node at (2, 2.4) {smaller $r$};
 \node at (2, 2) {desirable};
 
 \draw [thick, dashed] (5.0, .25) -- (5., 1.5);
 \draw [thick, dashed] (6.50, .25) -- (6.5, 1.5);
 \draw [thick,decorate,decoration={brace,amplitude=2pt},xshift=0.4pt,yshift=-0.4pt](5,1.55) -- (6.5,1.55) node[black,midway,yshift=-0.6cm] {};
 \node at (5.75, 2.4) {larger $r$};
 \node at (5.75, 2) {desirable};
 
 \draw [thick, ->] (0, -2.5) -- (7.5, -2.5); 
 \node at (3.75, -3) {$t$};
 
 \draw [thick, blue] (0.1,-1.5) -- (2, -1.5);
 \draw [thick, blue] plot [smooth] coordinates {(2, -1.5) (2.15, -2.25) (2.7, -1.6) (3, -1.5) (3.2, -1.5) };
 \draw [thick, blue] (3.2,-1.5) -- (7.5, -1.5);

 \draw [thick, dashed] (1.50, -.95) -- (1.5, -2.25);
 \draw [thick, dashed] (2.50, -.95) -- (2.5, -2.25);
  \draw [thick, dashed] (5.0, -.95) -- (5., -2.25);
 \draw [thick, dashed] (6.50, -.95) -- (6.5, -2.25);
 \draw [thick, ->] (0,-2.5) -- (0, -0.75);
 \node at (-.5, -1.625) {$r_t$};

\end{tikzpicture}
\caption{
Top: Measurements of a  non-stationarity univariate random variable, $X_t$ are shown
in grey together with the true mean in blue. 
This figure serves to highlight how the optimal choice of a forgetting factor or window length
may depend on location within a dataset. 
It 
follows that in the proximity of the change-point we wish $r$ to be small in order for it to 
adapt to change quickly. However, when the data is itself 
piece-wise stationary, we wish for $r$ to be large in order to be able 
to fully exploit all relevant data.\\
Bottom: An illustration of how an ideal adaptive forgetting factor would behave; decreasing directly after 
a change occurs and quickly recovering thereafter.
}
\label{h_size}
\end{figure}
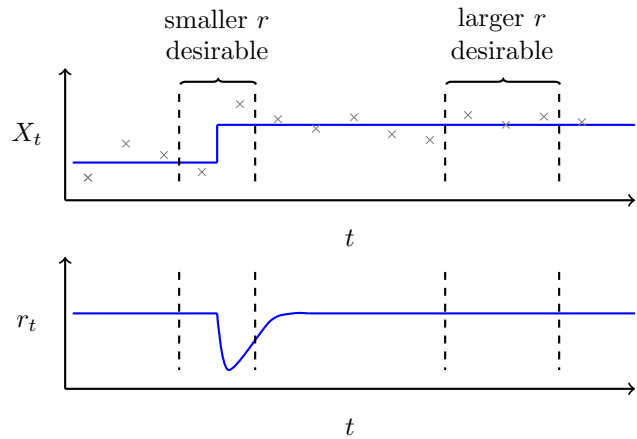

However, it is important to note that for any non-stationary data the optimal choice of these parameters may depend on 
the location within the dataset. By this we mean that in the proximity to a change-point it would clearly be desirable to
have smaller choice of $h$ and $r$; thereby reducing the influence of old, irrelevant observations.
Whereas within a locally stationary region we wish to have a larger choices of $h$ and $r$ in order to 
effectively learn from a wide range of pertinent observations. This concept is
demonstrated pictorially in the top panel of Figure [\ref{h_size}].

In the case of real-time fMRI and neurofeedback we inherently expect the
statistical properties of a subject's
functional connectivity networks to vary depending on both the task at hand as well as the neurofeedback. Therefore,
the choice of a fixed window length, $h$, or forgetting factor, $r$, may be inappropriate. 

In order to address this issue we propose the use of an adaptive forgetting methodology
\citep{haykin2008adaptive}. 
This corresponds to a selection of methods where the magnitude of the forgetting factor
is adjusted directly from 
the data in real-time. As a result, the value of the forgetting factor has a direct dependence on the 
time index, $t$. To make this relationship explicit we write $r_t$ to denote the adaptive forgetting 
factor at time $t$. The 
bottom panel of Figure [\ref{h_size}] provides an 
illustration of desirable behaviour for an adaptive forgetting factor. We note that 
immediately after a change occurs the forgetting factor drops. This helps
discard past information which is no longer relevant and gives additional weighting to 
new observations. Moreover, it is also important to note that in the presence of piece-wise stationarity data the 
value of the adaptive forgetting factor increases, allowing for a larger number of observations to be leveraged and yielding 
more accurate and stable estimates. 

Moreover, the use of adaptive forgetting
also provides an additional monitoring mechanism. 
By considering the estimated value of the forgetting factor $r_t$
at any given point in time we can gain an understanding as to the current degree of 
non-stationarity in the data \citep{anagnostopoulos2012online}. 
This follows from the fact that the estimated forgetting factor quantifies the influence of
recent observations on the sample mean and covariance. Thus it follows that large values of $r_t$ are
indicative of piece-wise stationarity whereas small values of $r_t$ provide evidence for changes in the network structure.

In order to effectively learn the forgetting factor in real-time we 
require a data-driven approach. One popular solution is to
empirically measure performance of current estimates by 
calculating the likelihood of incoming observations. 
In this way we are able to measure the 
performance of an estimated mean, $\bar{x}_t$, and sample covariance,  $S_t$, when provided with 
unseen data; thereby providing the basis on which to update our choice of forgetting factor.
Under the assumption that all observations follow a 
multivariate Gaussian distribution, this likelihood of a new observation $X_{t+1}$ is:
\begin{equation}
 \mathcal{L}_{t+1} = \mathcal{L} (X_{t+1}; \bar x_t, S_t) = - \frac{1}{2} \mbox{ log det} (S_t) - \frac{1}{2}(X_{t+1} - \bar x_t)^T S^{-1}_t (X_{t+1} - \bar x_t).
\end{equation}
While it would be possible to maximize 
$\mathcal{L}_{t+1}$ 
using a cross-validation framework in 
an offline setting, such an approach is challenging in a real-time setting.
This is because cross-validation approaches typically 
consider general performance over many subsets of past observations; 
therefore incurring a 
high computational cost. Moreover due to the highly autocorrelated nature of
fMRI time series, splitting past observations into subsets is itself non-trivial.  
Here we build on the work of \cite{anagnostopoulos2012online} and employ 
adaptive forgetting methods to maximize this quantity in a  computationally efficient manner. This is achieved by approximating the 
derivative of $\mathcal{L}_{t+1}$ with respect to $r_t$. This derivative can subsequently be used to update $r_t$ in a 
stochastic gradient ascent framework \citep{bottou2004stochastic}.

From equations (\ref{EWMAmu}), (\ref{EWMApi}) and (\ref{EWMAcov}) we can see the direct dependence of estimates $\bar x_t$ and $S_t$
on a fixed forgetting factor $r$.
This suggests that the likelihood is itself a function of the forgetting factor, allowing us to 
calculate its derivative with respect to $r$ as follows:
\begin{align}
 \mathcal{L}_{t+1}' &= \frac{ \partial \mathcal{L}_{t+1} }{\partial r}\\
 \label{likelihoodDerivative}
  &= \frac{1}{2}(X_{t+1} - \bar x_t)^T \left ( 2 S_t^{-1} \bar x_t' -  S_t^{-1} S_t' S_t^{-1} (X_{t+1} - \bar x_t) \right ) - \frac{1}{2} \mbox{trace}~ (S_t^{-1} S_t')
\end{align}
where have written $A'$ to denote the derivative of $A$ with respect to $r$ (i.e.,  $\frac{\partial A}{\partial r}$). Full details 
are provided in Appendix A. 

Given the derivative, $\mathcal{L}_{t+1}'$, we can subsequently update our choice of forgetting factor using gradient ascent:
\begin{equation}
\label{AFupdate}
 r_{t+1} = r_t + \eta \mathcal{L}_{t+1}',
\end{equation}
where $\eta$ is a small step-size parameter. Equation (\ref{AFupdate}) serves to highlight the strengths of adaptive forgetting;
by calculating
$\mathcal{L}_{t+1}'$ we are able to learn the direction along $r_t$ which maximizes the log-likelihood of unseen observations. It follows that if 
$\mathcal{L}_{t+1}'$ is positive, $r_t$ should be increased, while the converse is true if $\mathcal{L}_{t+1}'$  is negative. Moreover,
in calculating $\mathcal{L}_{t+1}'$ we also learn a magnitude.
This implies that all updates in equation (\ref{AFupdate}) will be of a different
order of magnitude. 
This is fundamental as it allows for rapid adjustments in the presence of abrupt changes. 

Finally, once $r_{t+1}$ has been calculated, we are able to learn estimates $\bar x_{t+1}$ and $S_{t+1}$ using the same
recursive equations (\ref{EWMAmu}) - (\ref{EWMAcov}) with the minor amendment that the effective sample
size, $\omega_t$ is calculated as:
\begin{equation}
\label{EWMAomegaAdaptive}
\omega_t =  r_{t-1} \cdot \omega_{t-1} + 1.  
\end{equation}

\subsection{Real-time network estimation}
\label{sec--rtSINGLEAlgo}

We now turn to the problem of estimating the precision 
matrix at time $t$.
In this section, we describe how we can adapt the SINGLE algorithm in such a manner that we can obtain 
an estimated precision matrix that is both sparse and temporally homogeneous in real time.

Given a sequence of estimated sample covariance matrices $\{S_t\} = \{S_1, \ldots, S_T\}$, the SINGLE algorithm 
is able to estimate corresponding precision matrices,
$\{ \Theta_t \} = \{ \Theta_1, \ldots, \Theta_T\}$,
by solving the following convex optimization problem:
\begin{equation}
\label{SINGLEobj}
\{  \Theta_t \} = \underset{\{ \Theta_t \}}{\mbox{argmin}} \left \{ \sum_{i=1}^T - \mbox{log det }  \Theta_i + \mbox{trace } (S_i \Theta_i) + \lambda_1 \sum_{i=1}^T||\Theta_i||_1 + \lambda_2 \sum_{i=2}^T||\Theta_i - \Theta_{i-1}||_1 \right \}.
\end{equation}
The first sum in equation (\ref{SINGLEobj}) 
corresponds to a likelihood term while the remaining 
terms, parameterized by $\lambda_1$ and $\lambda_2$ respectively, enforce sparsity and temporal 
homogeneity constraints. Estimated precision matrices, $\{  \Theta_t \}$, therefore balance a trade-off between
adequately describing observed data and satisfying sparsity and temporal homogeneity constraints.

However, in the real-time setting, a new $S_t$ is constantly obtained implying that the dimension of the solution to equation (\ref{SINGLEobj})
grows over time. It follows that iteratively re-solving equation (\ref{SINGLEobj}) is both wasteful and
computationally expensive. In particular, valuable computational resources will be spent estimating
past networks which are no longer of interest. In order to address this issue the following objective function is proposed to estimate
the functional connectivity network at time $t$:
\begin{equation}
\label{rtSINGLEobj}
f(\Theta) = - \mbox{log det }  \Theta + \mbox{trace } (S_t \Theta) + \lambda_1 ||\Theta||_1 + \lambda_2 ||\Theta - \Theta_{t-1}||_1,
\end{equation}
where $\Theta_{t-1}$ 
corresponds to the estimate of the precision matrix at time $t-1$ and
is assumed to be fixed.
The proposed real-time SINGLE (rt-SINGLE) algorithm is thus able to accurately estimate 
$\Theta_t$ by minimizing equation (\ref{rtSINGLEobj}) --- in doing so the proposed method 
must find a balance between goodness-of-fit and satisfying the conditions of sparsity and temporal homogeneity. 
The former is captured by the likelihood term:
\begin{equation}
\label{likeTerm}
 l(\Theta) = - \mbox{log det }  \Theta + \mbox{trace } (S_t \Theta),
\end{equation}
and provides a measure of how precisely  $\Theta$ describes the 
current estimate of the sample covariance, $S_t$. The latter two terms of the objective correspond to regularization
penalty terms:
\begin{equation}
\label{penTerm}
g_{\lambda_1, \lambda_2} = \lambda_1 ||\Theta||_1 + \lambda_2 ||\Theta - \Theta_{t-1}||_1
\end{equation}
The first of these, parameterized by $\lambda_1$, encourages sparsity while the second, parameterized by $\lambda_2$,
determines the extent of temporal homogeneity. 
By penalizing changes in functional connectivity networks, the second penalty encourages 
sparse innovations in edge structure over time. As a result, network changes are only reported when heavily substantiated by evidence in 
the data. It is also important to
note that equations (\ref{likeTerm}) and (\ref{penTerm}) serve to formalize the separable nature of 
our objective function; a property we use to our advantage in the optimization algorithm.

\subsubsection{Optimization algorithm}
\label{rtSINGLEalgoDescription}

In order to efficiently minimize the rt-SINGLE objective function we introduce further adjustments.
Equations (\ref{rtSINGLEobj})-(\ref{penTerm}) clearly expose the separable nature of the objective, which can 
be expressed as the sum of two sub-functions. It is precisely this property which is exploited in the
original SINGLE algorithm by employing an Alternating Directions Method of Multipliers (ADMM) algorithm \citep{ADMM}. 
The ADMM is a form of augmented Lagrangian algorithm that is particularly well suited to addressing this class 
of separable and 
highly structured minimization problems. Formally, such an algorithm proceeds by iteratively minimizing each 
of the sub-functions together with 
an additional Lagrangian penalty term. As we demonstrate below, each of these minimization problems will 
either have a closed form solution or
can be efficiently solved.

As in the SINGLE algorithm, we proceed by introducing an auxiliary variable $Z \in \mathbb{R}^{p \times p}$. 
Here $Z$ corresponds directly 
to $\Theta$ and we require $Z = \Theta$ for convergence. Minimizing equation (\ref{rtSINGLEobj}) can subsequently
be cast as the following constrained optimization problem:
\begin{align}
\label{sep1}
 \underset{\Theta, Z}{\mbox{minimize}} \hspace{5mm} &  \left \{  -\mbox{log det } \Theta + \mbox{ trace } ( S_t \Theta)  + \lambda_1 ||Z||_1 + \lambda_2  ||Z - \Theta_{t-1}||_1 \right  \} \\
 \label{sep2}
 \mbox{subject to} \hspace{5mm} & \Theta = Z.
\end{align}
We note that $\Theta$ is now only involved in the likelihood component while $Z$ is involved exclusively in the penalty component. 
Thus, by introducing $Z$ we have decoupled the initial objective function --- allowing us to take advantage of the
individual structure associated with each term.

As in the SINGLE algorithm, we formulate the augmented Lagrangian corresponding to equations (\ref{sep1}) and (\ref{sep2}), which
is defined as:
\begin{align}
\begin{split}
\label{aug_lagrange1}
\mathcal{L}_{\gamma} \left ( \Theta, Z, U \right ) = &-  \mbox{log det } \Theta + \mbox{ trace } (  S_t \Theta)  + \lambda_1 ||Z||_1 \\
&+\lambda_2 ||Z - \Theta_{t-1}||_1 +\nicefrac{1}{2}  \left (|| \Theta - Z + U ||_2^2 - || U||_2^2 \right ),
\end{split}
\end{align}
where $U \in \mathbb{R}^{p \times p}$ is the (scaled) Lagrange multiplier. Equation (\ref{aug_lagrange1}) corresponds to the
Lagrangian together with an additional quadratic penalty term which serves to both increase the 
robustness of the proposed method \citep{Bertsekas1982} as well as greatly simplify the resulting computations, as we describe below. 

The proposed estimation algorithm works by iteratively minimizing equation (\ref{aug_lagrange1}) with respect to $\Theta$ and $ Z$
while maintaining all other variables fixed.
In this way, we are able to decouple the augmented Lagrangian and exploit the individual structure corresponding to each of these variables. 
Due to the iterative nature of the algorithm, in what follows we write $\Theta^i$ to denote the estimate of $\Theta$ at the 
$i$th iteration. The same notation is used for both $Z$ and $U$. The algorithm is initialized with
$\Theta^i = I_p$, $Z^i = U^i = \mathbf{0} \in \mathbb{R}^{p \times p}$. 
We note that the $\Theta$ and $U$ update steps remain unchanged from the original offline algorithm.
However, in the case of the $Z$ update an adjustment is required due to the fact that 
past networks, $\Theta_{t-1}$, are treated as constants. 
Subsequently, $Z$ is updated by solving:
\begin{equation}
\label{step22}
 Z^i = \underset{Z}{\mbox{argmin}} \hspace{2.5mm}  \left \{
  \nicefrac{1}{2}|| \Theta^i - Z + U^{i-1} ||_{2}^2 + \lambda_1 ||Z||_1 +\lambda_2 ||Z - \Theta_{t-1}||_1  \right \},
\end{equation}
where $\Theta^i, U^i$ and $\Theta_{t-1}$ are treated as constants.
Here we note that equation (\ref{step22}) involves a series of one-dimensional problems as only element-wise operations are applied. 
This implies that we may solve an independent problem of the following form for each entry in $Z$:
\begin{equation}
\label{fused_objective_thing}
 \underset{(Z)_{k,l} \in \mathbb{R}}{\mbox{argmin}} \hspace{2.5mm}  \left \{   \nicefrac{1}{2} || (\Theta^i - Z + U^{i-1})_{k,l}||_{2}^2 
+ \lambda_1 ||(Z)_{k,l}||_1 +\lambda_2 ||(Z - \Theta_{t-1})_{k,l}||_1 \right \}
\end{equation}
\normalsize
where we write $(M)_{k,l}$ to denote the $(k,l)$ entry for any square matrix $M$.
Thus each element of $Z$ can be updated by solving a one-dimensional convex problem. While there is no closed form solution,
we may employ efficient line search algorithms \citep{BoydBook, Nocedal2006}.
Due to the symmetric nature of $Z$ it follows that only $\frac{p(p+1)}{2}$ of such problems must be solved.

\subsubsection{Burn-in period}
\label{sec--BurnIn}
It is common for real-time algorithms to incorporate a brief burn-in phase a when they are initialized.
This involves collecting the first $N_{{BurnIn}}$ observations and using these to collectively obtain the first estimate. 
Many times such an approach is motivated by the need to ensure sample statistics are well-defined, however
due to the presence of regularization the proposed method does not require a burn-in \textit{per se}. 
That said the use of a burn-in phase can improve initial network estimates and 
thereby result in improved network estimation overall. 
As a result, the first $N_{{BurnIn}}$ observations are collected and 
used to estimate the corresponding precision matrices by directly
applying the offline SINGLE algorithm. This involves solving
equation (\ref{SINGLEobj}). From then onward, new estimates of
the precision matrix are obtained as described previously.

\begin{algorithm}[h!]
 \SetAlgoLined
 
  \textbf{Input:} New observation $X_t$ as well as previous estimates $\bar x_{t-1}$, $\omega_{t-1}$, $\Pi_{t-1}$.\\
  Fixed forgetting factor
  $r \in (0,1]$ or stepsize parameter $\eta$, \\
  penalty parameters $\lambda_1, \lambda_2$, 
  convergence tolerance $\epsilon$, \\
 \KwResult{Sparse estimates of precision matrix $\Theta_t$} 
 \#\# Update forgetting factor\;
 $r_t = r_{t-1} + \eta \mathcal{L}_t'$ \# note that $r_t = r ~\forall t $ in the case of EWMA models\;
 \#\# Update $\omega_t, \bar x_t$ and $S_t$\;
$ \omega_t = r_t \cdot \omega_{t-1} + 1$\;
$ \bar x_t = (1 - \frac{1}{\omega_t}) \cdot \bar x_{t-1} + \frac{1}{\omega_t} \cdot X_t$\;
$ \Pi_t = (1 - \frac{1}{\omega_t}) \cdot  \Pi_{t-1} + \frac{1}{\omega_t} X_t X_t^T $\;
$ S_t = \Pi_t - \bar x_t \bar x_t^T$\;

\#\# Begin optimization algorithm\;
$\Theta_t^0 = I_p, Z_t^0 = U_t^0 = \mathbf{0}_p$\;
Convergence = False\;
  \While{Convergence==False}{

  \#\# $\Theta_t$ Update\;
  $V,D = \mbox{eigen} \left (S_t -  \left (Z_t^{i-1} - U_t^{i-1} \right ) \right )$\;
  $\tilde D = \mbox{diag} \left ( \frac{1}{2} \left ( -D + \sqrt{D^2 + 4} \right )\right )$\;
  $\Theta_t^i = V \tilde D V'$\;
  
  \#\# $Z_t$ Update\;
  \For{ \mbox{each} $l,k$}{
  $(Z_t)_{l,k} = \underset{ x \in \mathbb{R} }{\mbox{argmin}} \left \{  \nicefrac{1}{2} ((\Theta_t^i+U_t^{i-1})_{k,l}-x)^2 + \lambda_1 ||x||_1 + \lambda_2 ||x-(\Theta_{t-1})_{k,l}||_1 \right \} $
  }
  
  \#\# $\{U\}$ Update\;
  $U_t^i = U_t^{i-1} + \Theta_t^i - Z_t^i$\;
 
     \If{$||\Theta_t^i - Z_y^i||^2_2 < \epsilon \mbox{ and } ||Z_t^{i}-Z_t^{i-1}||_2^2 < \epsilon$ }{
   $\mbox{Convergence=True}$\;
   }
  }

\Return  $ \Theta_t$
 \caption{real-time SINGLE algorithm}
 \label{psuedo}
\end{algorithm}

\subsection{Parameter tuning}
\label{sec--rtSINGLEParams}

Parameter estimation is challenging in the real-time setting. 
Approaches such as cross-validation, which are inherently difficult to implement due to the non-stationarity
of the data, are further hampered by the limited computational resources. 
As an alternative, information theoretic
approaches such as minimizing the AIC or BIC may be taken but these too may incur a high computational burden. 
In this section we discuss the three parameters required in the proposed method
and provide a clear interpretation as well
as a general overview on how each should be set.

The use of a sliding window or EWMA model
implies an assumption about the non-stationarity of the data. Specifically, the underlying assumption behind such 
approaches is one of local, as opposed to global, stationarity. That is to say, we expect network
structure to remain constant within a neighbourhood
or any observation but to vary over a larger period of time. In choosing window length, $h$, or the fixed forgetting 
factor, $r$, we are inherently quantifying the size of this neighbourhood --- a small value of $h$ or $r$ implies a small
neighbourhood and is indicative of a system that varies quickly, while larger values imply slower, more gradual changes. 

Defined in equation (\ref{EWMAomega}), $\omega_t$ provides an estimate for the effective sample size. That is, $\omega_t$
is indicative of the number of observations used in the calculation of the mean and sample covariance respectively. We note that as 
$t$ becomes large we have:
\begin{equation}
\label{omegaChoice}
 \omega_t = \sum_{i=1}^t r^{t-i} \approx  \frac{1}{1-r}.
\end{equation}
This allows us to directly quantify the effect of $r$ on the number of observations employed in each calculation. 
Furthermore, equation (\ref{omegaChoice}) also provides a clear relationship between the choice of window length $h$ and
forgetting factor $r$.  

In practice, it is possible to choose the window length, $h$, or forgetting factor, $r$
based on prior belief regarding the degree of non-stationarity in the data or 
using a 
maximum likelihood framework \citep{lindquist2007modeling, lindquist2014evaluating}. 
A more elegant solution is provided via the use of adaptive forgetting methods. These methods designate
the choice of $r_t$ to the data. As a result, only the stepsize parameter, $\eta$, must be specified.
Typical choices of $\eta$ range from $0.001$ to $0.05$.

Parameters $\lambda_1$ and $\lambda_2$ enforce sparsity and temporal homogeneity respectively. The choice of these parameters 
affects the degrees of freedom of estimated networks, suggesting the use of information theoretic approaches such as AIC. 
However, in a real-time setting, choosing $\lambda_1$ and $\lambda_2$ in such a manner presents a computational burden. 
As a result, we propose two heuristics for choosing appropriate values of $\lambda_1$ and $\lambda_2$ respectively. 
One potential approach involves studying a previous scan of the subject in question. If this is available then
the regularization parameters may be chosen by minimizing AIC over this scan. Alternatively, the burn-in phase may be
used to choose adequate parameters. Such an approach would involve choosing $\lambda_1$ and $\lambda_2$ which minimized AIC 
over the burn in period. 
Moreover, it is worth noting that tuning $\lambda_1$ and $\lambda_2$ adaptively in a similar manner to the forgetting
factor presents theoretical and computational challenges 
due to the non-differentiable nature of the regularization penalties.

\section{Simulation study}
\label{sec--rtSINGLEsims}
\subsection{Simulation settings}

In this section we evaluate the performance of the rt-SINGLE algorithm through a series of simulation studies.
In each simulation we produce simulated time series data giving rise to a number of 
connectivity patterns which reflect those reported in real fMRI data. The objective is then to measure
whether our proposed algorithm is able recover the underlying patterns in real-time.
We are primarily interested in studying the performance of the proposed methods in two
ways; first we wish to study the quality of the estimated covariance matrices over time. That is to say,
we study how accurately our sample covariances represent the true underlying covariance structure.
Second, we are also interested in the correct estimation of the presence or absence of edges.

There are two main properties of fMRI data which we wish to recreate in the simulation study. The first is the high autocorrelation
which is typically present in fMRI data \citep{handbook}. The second and main property we wish to recreate
is the structure of the connectivity networks themselves. It is widely reported that brain networks have a small-world topology as well as 
highly connected hub nodes \citep{bullmore} and we therefore look to enforce these properties in our simulated data.

Vector Autoregressive (VAR) processes are well suited to the task of producing autocorrelated multivariate time series as they are capable of
encoding 
autocorrelations within components as well as cross correlations
across components \citep{DCR}. 
The focus of these simulations is to study the performance of the proposed method 
in the presence of non-stationary data. As a result the simulated datasets are only locally stationary. This is achieved by concatenating
multiple VAR process which are simulated independently --- this results in abrupt changes which are representative of 
the typical block structure of task based fMRI experiments.

Moreover, when simulating connectivity structures we study the performance of the proposed algorithm using two types of random graphs;
scale-free random graphs obtained by using
the preferential attachment model of \cite{barabasi1999emergence}
and small-world random graphs obtained using the \cite{watts1998collective} model.
The use of each of these types of networks is motivated by the fact that they are each known to each resemble different aspects of fMRI networks.
Throughout each of the simulations, first the network architecture was simulated using either of the aforementioned methods. Then
edge strength was uniformly sampled from 
$[-\nicefrac{1}{2},-\nicefrac{1}{4}] \cup [\nicefrac{1}{4}, \nicefrac{1}{2}]$.
This introduced further variability into the simulated networks, increasing the difficulty of each task.

The simulations presented in this work look to quantify the ability of the rt-SINGLE algorithm 
to accurately estimate time-varying networks in real-time. 
In simulation \rom{1} we study the quality of estimated covariance matrices over time. 
In simulations \rom{2} and \rom{3} we 
consider the overall performance of the proposed method by
generating connectivity structures according to scale-free and small-world networks respectively.
Finally, in Simulation \rom{4} we look to quantify the computational cost of the proposed method as 
the number of nodes, $p$, increases. This simulation is fundamental in the neurofeedback setting as 
subjects must receive prompt and accurate feedback.
Throughout this section we compare results for the rt-SINGLE algorithm where a fixed forgetting factor (corresponding to an EWMA model) is 
employed as well as adaptive forgetting techniques. Further,
we also consider the performance 
of the offline SINGLE algorithm as a benchmark. Naturally we expect the rt-SINGLE algorithms to generally 
perform below its offline counterpart, however, the difference in performance
will be indicative of how well the proposed methods work.

Throughout each of these simulations, the parameters for the offline SINGLE algorithm where determined as described 
in \cite{Monti2014}. That is, the choice of 
 kernel width was obtained by maximizing leave-one-out log-likelihood while
the choice of regularization parameters where chosen by minimizing AIC. 
In the case of the real-time algorithms the parameters where chosen as follows. 
The fixed forgetting factor was chosen to be $r=0.95$ as this corresponded approximately to 
an effective sample size of twenty observations. While in the case of adaptive forgetting
$\eta=0.005$ was chosen and a burn-in period of 15 observations was used.
Regularization parameters 
where chosen to minimize AIC over the burn-in period. 

\subsection{Performance measures}

As alluded to previously, we wish to evaluate the performance of the proposed method in two 
distinct ways. First, we wish to study
the reliability with which we can track changes in covariance structure using either 
a fixed forgetting factor or an adaptive forgetting factor. 
In order to quantify the difference between the true covariance structure 
and our estimated covariance we consider the distance defined by the trace inner product:
\begin{equation}
\label{TIP}
 d (\Sigma, S) = \mbox{Trace}~(\Sigma^{-1} S).
\end{equation}
It follows that if the estimated sample covariance, $S$, is a good estimate of the true
covariance, $\Sigma$, we will have that $d(\Sigma, S) \approx p$. However, if $S$ is a poor estimate,
the distance $d$ will be large. Moreover, since both $\Sigma$ and $S$ are positive definite we have that
$d(\Sigma, S)$ will always be positive. 

Second, we wish to consider the estimated functional connectivity networks at each point in time.
In this application we are particularly interested in
correctly identifying the non-zero entries in estimated precision
matrices, $\hat \Theta_i$, at each $i=1, \ldots, T$. An edge is assumed to be present between the $j$th and $k$th nodes 
if $( \hat \Theta_i)_{j,k} \neq 0$. At the $i$th observation we define the set of all reported edges as $D_i = \{ (j,k): (\hat \Theta_i)_{j,k} \neq 0\}$.
We define the corresponding set of true edges as $T_i= \{ (j,k): ( \Theta_i)_{j,k} \neq 0\}$ where we write $\Theta_i$ to denote the 
true precision matrix at the $i$th observation.
Given $D_i$ and $T_i$ we consider a number of
performance measures at each observation.

First we measure the precision, $P_i$. This measures the percentage of reported
edges which are actually present (i.e., true edges). Formally, the precision is given by:
$$ P_i =  \frac{|D_i \cap T_i|}{|D_i|}.$$
Second we also calculate 
the recall, $R_i$, formally defined as:
$$ R_i =  \frac{|D_i \cap T_i|}{|T_i|}. $$
This measures the percentage of true edges which were reported by each algorithm.
Ideally we would
like to have both precision and recall as close to one as possible. Finally, the $F_i$ \textit{score}, defined as
\begin{equation}
\label{f_eq}
F_i = 2  \frac{P_i  R_i}{P_i + R_i},
\end{equation}
summarizes both the precision and recall by taking their harmonic mean. It follows that $F_i$ will lie
on the interval $[0,1]$ with $F_i=1$ indicating perfect performance.

\subsubsection{Simulation \rom{1} --- Covariance tracking}

In this simulation we look to assess how accurately we are able to track changes in
covariance structure via the use of fixed (i.e., EWMA models) and adaptive 
forgetting factors. 
As discussed previously, obtaining accurate estimates of the sample
covariance in real-time is a fundamental problem both
when studying functional connectivity networks in general and in particular for the proposed methods. 

Here datasets were simulated as follows: each dataset consisted of five segments each of length 100 (i.e., overall duration of 500).
The network structure within each segment were simulated according to either to either the 
\cite{barabasi1999emergence} preferential attachment model or using the \cite{watts1998collective} model.
The use of each of these models was motivated by the fact that they are able to generate
scale-free and small-world networks respectively; two classes of networks which are frequently encountered in
the analysis of fMRI data \citep{eguiluz2005scale, bassett2006small, sporns2004organization}.

In this simulation the estimated sample covariances from the proposed methods were compared to 
the results when using a symmetric Gaussian kernel, as in the offline SINGLE algorithm.
The choice of kernel width was determined by maximizing leave-one-out log-likelihood. In the 
case of the fixed forgetting factor, $r=0.95$ was chosen as this 
corresponded to an effective sample size of twenty observations. Finally, in the case of adaptive forgetting
$\eta=0.005$ was chosen.

\begin{figure}[ht]
\centering
\includegraphics[width=\textwidth]{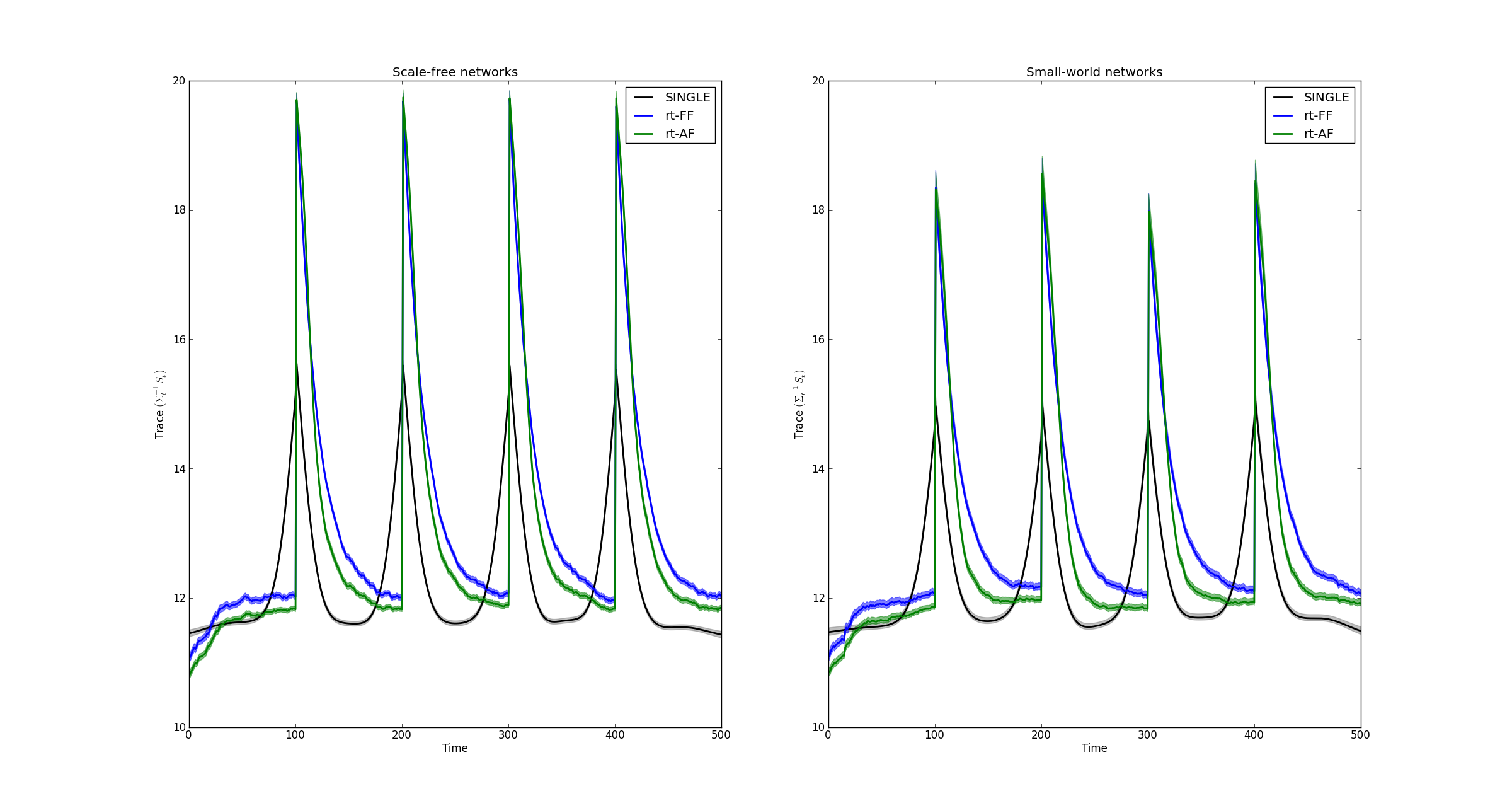}
\caption{
In this simulation we study the capability of the proposed algorithm to accurately track changes 
to covariance structure over time. In order to quantify this we 
consider the distance defined by the trace inner product, given in equation (\ref{TIP}). \\
Left: Covariance tracking results when underling network structure is
simulated according to the scale-free preferential attachment model of \citep{barabasi1999emergence}.
A change occurred every 100 observations. We note that the symmetric Gaussian kernel employed for the offline 
SINGLE algorithm outperforms the online algorithms as expected. However, when the 
covariance structure remains piece-stationary for extended periods of time the online algorithms
are able to outperform their offline counterparts.\\
Right: Covariance tracking results when the 
underlying network structure was simulated using
small-world random networks according to the model of \cite{watts1998collective}.}
\label{simCovTrack}
\end{figure}

Figure [\ref{simCovTrack}] shows results when scale-free (left) and small-world (right) network structures are simulated.
We note that 
the quality of the estimated covariances drops in the proximity of a change-point for all three 
algorithms. 
In the case of the offline SINGLE algorithm this drop is symmetric due to
the symmetric nature of the Gaussian kernel employed. However, in the case of the real-time algorithms 
the drop is highly asymmetric and occurs directly after the change-point, as is to be expected. 
Due to the sudden change in covariance structure, these methods suffer immediately after abrupt changes
in covariance structure, but are able to quickly recover.
Moreover, from Figure [\ref{simCovTrack}] we note that the covariance tracking capabilities of the proposed methods are 
not adversely affected by the choice of underling network structure.

\subsubsection{Simulation \rom{2} --- Scale-free networks}

In this simulation we look to 
obtain a general comparison between the rt-SINGLE algorithm and its offline counterpart. 
We simulated datasets with the following structure: each dataset consisted of five segments each of length 100 (i.e., overall duration of 500).
The network structure within each segment was independently simulated according to the \cite{barabasi1999emergence} preferential 
attachment model.
The motivation behind the use of the \cite{barabasi1999emergence} model is based on evidence that 
brain networks are scale-free, implying that the degree distribution follows a power law. This implies the 
presence of a reduced number of \textit{hub} nodes which have access to many other regions, while the remaining majority of nodes 
have a small number of edges \citep{eguiluz2005scale}.

In this simulation the entire dataset was simulated \textit{apriori}. In the case of the rt-SINGLE algorithms, one observation was 
provided at time, thereby treating the dataset as if it was a stream arriving in real-time. 
The offline SINGLE algorithm was provided with the entire dataset and this was treated as an offline task.

\begin{figure}[ht]
\centering
\includegraphics[width=\textwidth]{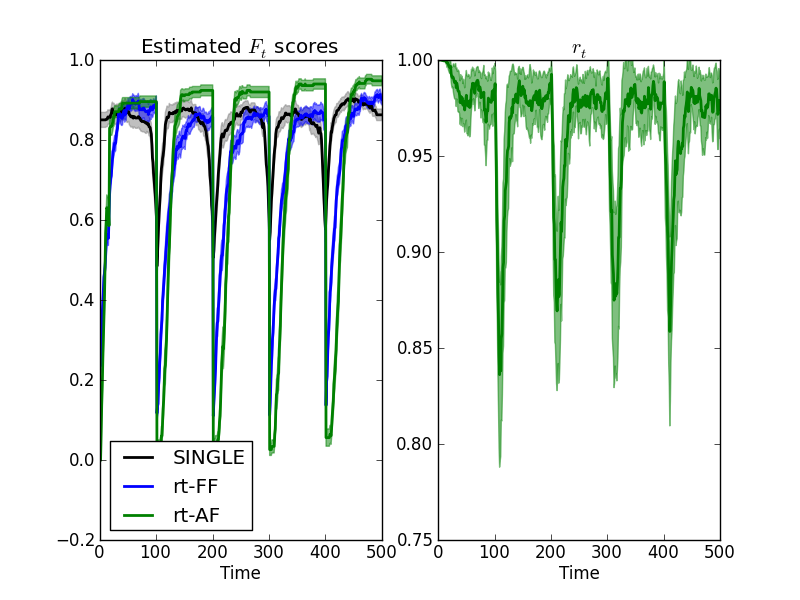}
\caption{
Left: 
Mean $F$ scores for the offline SINGLE algorithm and the real-time algorithms employing a 
fixed forgetting factor (rt-FF) and adaptive forgetting respectively (rt-AF).
Here the underlying network structure was simulated using
scale-free random networks according to the preferential attachment model of \cite{barabasi1999emergence}. 
A change occurred every 100 time points. We note that all three algorithms experience a drop
in performance in the vicinity of these change-points, however in the case of the real-time algorithms the drop is asymmetric. 
Moreover, we further note that when adaptive forgetting is employed the real-time algorithm is able to outperform its
offline counterpart in sections where the data remains piece-wise stationary for long periods of time.\\
Right: mean values for the estimated adaptive forgetting factor, $r_t$, over time. We note there is a sudden drop directly
after changes occurs allowing the algorithm to adequately discard irrelevant information. }
\label{simSF}
\end{figure}

In the left panel of Figure [\ref{simSF}] we see the average $F_t$ scores for each of the real-time algorithms as well as the offline algorithm over 
100 simulations. We note that all three algorithms experience a drop in $F$-score in the proximity of change-points. 
The offline SINGLE algorithm is based on a symmetric Gaussian kernel,
as a result, we note that there it has a symmetric drop in performance in the vicinity of a change-point before quickly recovering. 
Alternatively, the drop in performance of the rt-SINGLE algorithms is asymmetric. This is due to the real-time nature of these algorithms. 
Moreover, we note that while the rt-SINGLE algorithm performs worse than its offline counterpart directly after
change-points, it is able to quickly recover to the level of the offline SINGLE algorithm. Specifically, in the case where 
adaptive forgetting is used, the real-time algorithm is able to outperform its offline counterpart 
in sections where the data remains piece-wise stationary for long periods of time.
This is
because it is able to increase the value of the adaptive forgetting factor accordingly. This allows the algorithm
to exploit a larger pool of relevant information compared to its offline counterpart. 
This is demonstrated on the right panel of Figure [\ref{simSF}] where the mean value of the 
adaptive forgetting factor is plotted. We see there is a drop directly after changes occur; this allows the
algorithm to quickly forget past information which is no longer relevant. We also note that the estimate value of the 
forgetting factor increases quickly after changes occur.

\subsubsection{Simulation \rom{3} --- Small-world networks}

While in Simulation \rom{2} scale-free networks were studied, it has been reported that brain networks follow 
a small-world topology \citep{bassett2006small}. Such networks are characterized by their high clustering coefficients which 
has been reported in both anatomical as well as functional brain networks \citep{sporns2004organization}. 

The \cite{watts1998collective} model works as follows:
starting with a regular lattice, the model is parameterized by $\beta \in [0,1]$ which quantifies the probability of randomly 
rewiring an edge. It follows that setting $\beta=0$ results in a regular lattice, while setting $\beta=1$ results in an 
 Erd\H{o}s-R\'{e}nyi (i.e., completely random)
network structure.
Throughout this simulation we set $\beta=\nicefrac{3}{4}$ as this yielded networks with sufficient variability but which still
displayed the desired small-world properties.

As in Simulation \rom{2}
the entire dataset was simulated \textit{apriori}. 
In the case of the rt-SINGLE algorithms, one observation was 
provided at a time, thereby treating the dataset as if it were arriving in real-time. 
In the case of the offline SINGLE algorithm, the algorithm was provided with the entire dataset.

\begin{figure}[ht]
\centering
\includegraphics[width=\textwidth]{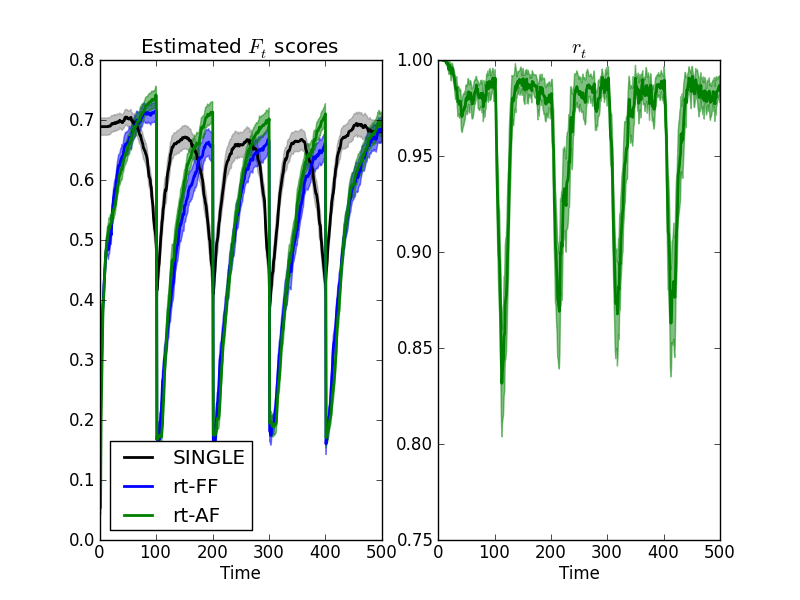}
\caption{
Left:
Mean $F$ scores for the offline SINGLE algorithm and the real-time algorithms employing a 
fixed forgetting factor (rt-FF) and adaptive forgetting respectively (rt-AF).
Here the underlying network structure was simulated using
small-world random networks according to the model of \cite{watts1998collective}. 
A change occurred every 100 time points. We note that all three algorithms experience a drop
in performance in the vicinity of these change-points, however in the case of the rt-SINGLE algorithms the drop is asymmetric.
Moreover, we further note that when adaptive forgetting is employed the real-time algorithm is able to outperform its
offline counterpart in sections where the data remains piece-wise stationary for long periods of time.\\
Right: mean values for the estimated adaptive forgetting factor, $r_t$, over time. We note there is a sudden drop directly
after changes occurs allowing the algorithm to adequately discard irrelevant information.
}
\label{simSW}
\end{figure}

In the left panel of Figure [\ref{simSW}] we see the average $F_t$ scores for each of the real-time algorithms as well
as the offline SINGLE algorithm over 
100 simulations. 
Due to the increased complexity of small-world networks, we note that the performance drops 
compared to scale-free networks considered in Simulation \rom{2}.
We further note that the rate at which the real-time networks recover after a change-point is 
reduced. 
As with Simulation \rom{2}, we note that both of the real-time algorithms are able to reach the same level of performance
as their offline counterpart if given sufficient time. Moreover, in the case where adaptive forgetting is employed
we once again find that the 
performance of the real-time algorithm exceeds that of the offline algorithm when the data is remains piece-wise stationary for a 
sufficiently long period of time. 
In the right panel of Figure [\ref{simSW}] we see the estimated adaptive forgetting factor over each of the 100 simulations. 
Again, we see the drop in the value of the forgetting factor directly after change-points; allowing past information 
to be efficiently discarded.

\subsubsection{Simulation \rom{4} --- Computational cost}

A fundamental aspect of real-time algorithms is that they must be computationally efficient in order 
to be able to update parameter estimates in the limited time provided. 
The main computational cost of the rt-SINGLE algorithm 
is related to the eigendecomposition of the $\Theta$ update,
which has a complexity of $\mathcal{O}(p^3)$ \citep{Monti2014}.

In this simulation we look to empirically study the computational cost. In this manner, we 
are able to provide a rough guide as to the number of ROIs which can be employed in a real-time 
neurofeedback study while still reporting network estimates at every point in time.
This was achieved by measuring the mean running time of each 
update iteration of the rt-SINGLE algorithm for various numbers of ROIs, $p$.

\begin{figure}[ht]
\centering
\includegraphics[width=\textwidth]{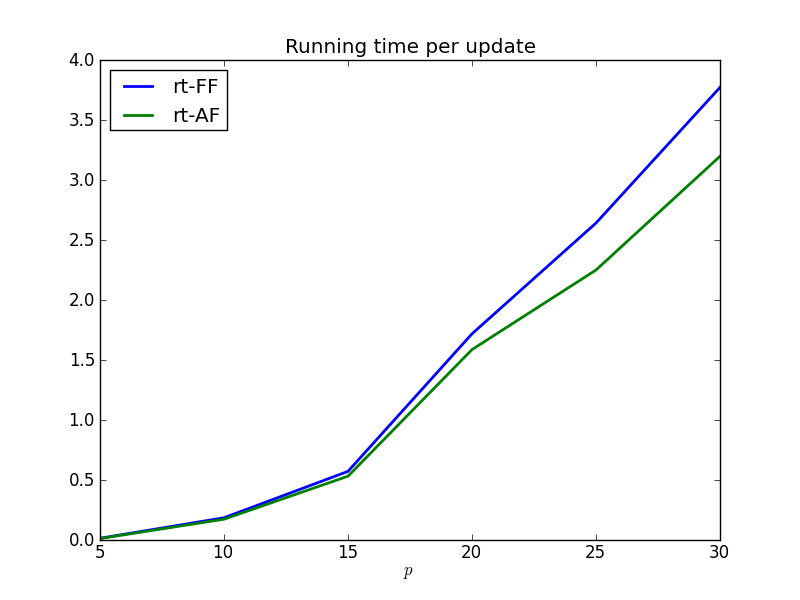}
\caption{
Mean running time (seconds) per update iteration of the rt-SINGLE algorithm when
either a fixed forgetting factor (rt-FF) or
adaptive forgetting (rt-AF) was employed. 
}
\label{simTime}
\end{figure}

Here each dataset was simulated as in Simulation \rom{2}; 
that is the underlying correlation was randomly generated according to a small-world
network.
However, here we choose to only simulate three segments, each of length 50, resulting in a 
dataset consisting of 150 observations.
For increasing values of $p$, the time taken to estimate a new precision matrix was calculated. 
Figure [\ref{simTime}] shows the mean running 
time for the rt-SINGLE algorithm where either
a fixed forgetting factor (i.e., an EWMA model) or adaptive forgetting was used. 
We note that the difference in computational cost between 
each of the algorithms is virtually indistinguishable. 

Finally we note that for $p<20$ nodes, it is possible to estimate functional connectivity 
networks in under two seconds,
making the proposed method practically feasible in many real-time studies. 
This simulation was run on a computer with an \textsc{Intel Core i5 CPU} at 2.8 GHz.

\section{Application: HCP motor-task fMRI data}
\label{sec--HCPapp}

In this section we present the first of our applications. 
Here 
a motor-task dataset from the Human Connectome Project \citep{elam2014human, van2012human} is studied.
While this dataset is not acquired and analyzed in real-time, it may be treated as such by only 
considering one observation at a time. 
This allows us to benchmark the rt-SINGLE algorithm to its offline counterpart using fMRI data as opposed to simulated 
examples, as we have done in Section \ref{sec--rtSINGLEsims}. 

\subsection{Motor-task data}
\label{data--HCP}
Twenty of the 500 available task-based fMRI datasets provided by the Human Connectome Project 
were selected at random.
Here subjects were asked to perform a simple motor task adapted from those developed by
\cite{buckner2011organization} and \cite{yeo2011organization}. 
This involved the presentation of visual cues asking subjects to
either tap their fingers (left or right), squeeze their toes (left 
or right) or move their tongue.
Each movement type was blocked, lasting 12 seconds, and was preceded by a three second visual cue. In addition there were
three 15 second fixation blocks per run\footnote{for further details 
please see http://www.humanconnectome.org/documentation/Q1/task-fMRI-protocol-details.html}. 

While this data is not intrinsically real-time --- in that the preprocessing was conducted after
data acquisition ---
it is included as a proof-of-concept study. 
The data was preprocessed offline as the focus lies on the comparison between the real-time 
and offline network estimation approaches rather than different preprocessing pipelines. Preprocessing 
involved regression of Friston’s 24 motion parameters and high-pass filtering using a cut-off frequency of $\nicefrac{1}{130}$Hz. 

Eleven bilateral cortical ROIs were defined
based on the Desikan-Killiany atlas \citep{desikan2006automated} covering 
occipital, parietal and temporal lobe (see Table \ref{table:MNI}).  The extracted 
time courses from these regions were subsequently used for the analysis. 
By treating the extracted time course data as if it was arriving in real-time 
(i.e., considering one observation at a time) we can compare the results of 
the proposed real-time method to offline algorithms while using the same underlying preprocessed data.

\begin{table}[ht]
\centering 
\begin{tabular}{l l l l l l l l} 
\hline \hline
 Name  &  \multicolumn{3}{c}{Right hem.} & & \multicolumn{3}{c}{Left hem.} \\ 
\hline 
 Lateral Occipital	 & 	31 & 	-84 & 	1 & & 	-29 & 	-87 & 	1\\
 Inferior Parietal & 43 &  -62 &  30 &      &  -39&  -68 &  30 \\
 Superior Parietal & 22 &  -62  & 48 &      &   -21&  -64 &  47 \\
 Precuneus &  11 &  -56 &  37 &      &  -10 &  -57 &  37 \\
 Fusiform &  34  &   -39 & -20  &      & -34 & -43 & -19 \\
 Lingual &  15  & -66 & -3 &      & -14 & -67 & -3 \\
 Inferior Temporal & 49 & -26 & -25 &      & -49 & -31 & -23 \\
 Middle Temporal & 57  & -22 & -14 &      & -56  & -27 &  -12 \\
 Precentral & 39 &-8 &43&      & -38 &-9& 43\\
 Postcentral & 42 & -21 & 44 & & -42 &-23& 44\\
 Paracentral & 9 &-26 &58 & & -8 & -28 &59 \\
\hline 
\end{tabular}
\caption{Regions and MNI coordinates} 
\label{table:MNI} 
\end{table}

\subsection{Results}

Both the SINGLE as well as the rt-SINGLE algorithms where applied to the motor-task fMRI dataset.
Our primary interest here is to report task-driven activations in functional connectivity.
In this way, we are able to examine if the rt-SINGLE algorithm is capable of reporting changes in 
functional connectivity resulting from changes in motor task. 

As a result, the functional relationships 
that were modulated by the motor task were studied; 
this corresponds to studying the edges in the estimated networks which are significantly correlated with
task onset.
This was achieved by first estimating time-varying functional connectivity networks using both the 
offline SINGLE algorithm as well as the proposed real-time algorithm.
In the case of the SINGLE algorithm, parameters where chosen as described in \cite{Monti2014}. 
This involved estimating the
width of the Gaussian kernel via leave-one-out cross validation and 
estimating regularization parameters via minimizing AIC. 
In the case of the real-time algorithm, adaptive forgetting was employed with $\eta=0.005$. 
The sparsity and temporal homogeneity parameters
where set to the same values as the offline SINGLE algorithm as the focus here was to study 
differences induced by estimating networks in real-time as opposed to differences resulting 
from different parameterizations.

The correlation between estimated functional relationships (i.e., edges) and the task-evoked 
HRF function were estimated using Spearman's 
rank correlation coefficient; a non-parametric measure of statistical dependence. 
The resulting $p$-values (one for each edge) were then corrected 
for multiple comparisons via the  Holm-Bonferroni method \citep{holm1979simple}.  This allowed us to obtain an activation network, 
summarizing which edges are statistically activated by task onset, for each algorithm.

\begin{figure}[ht]
\centering
\includegraphics[width=\textwidth]{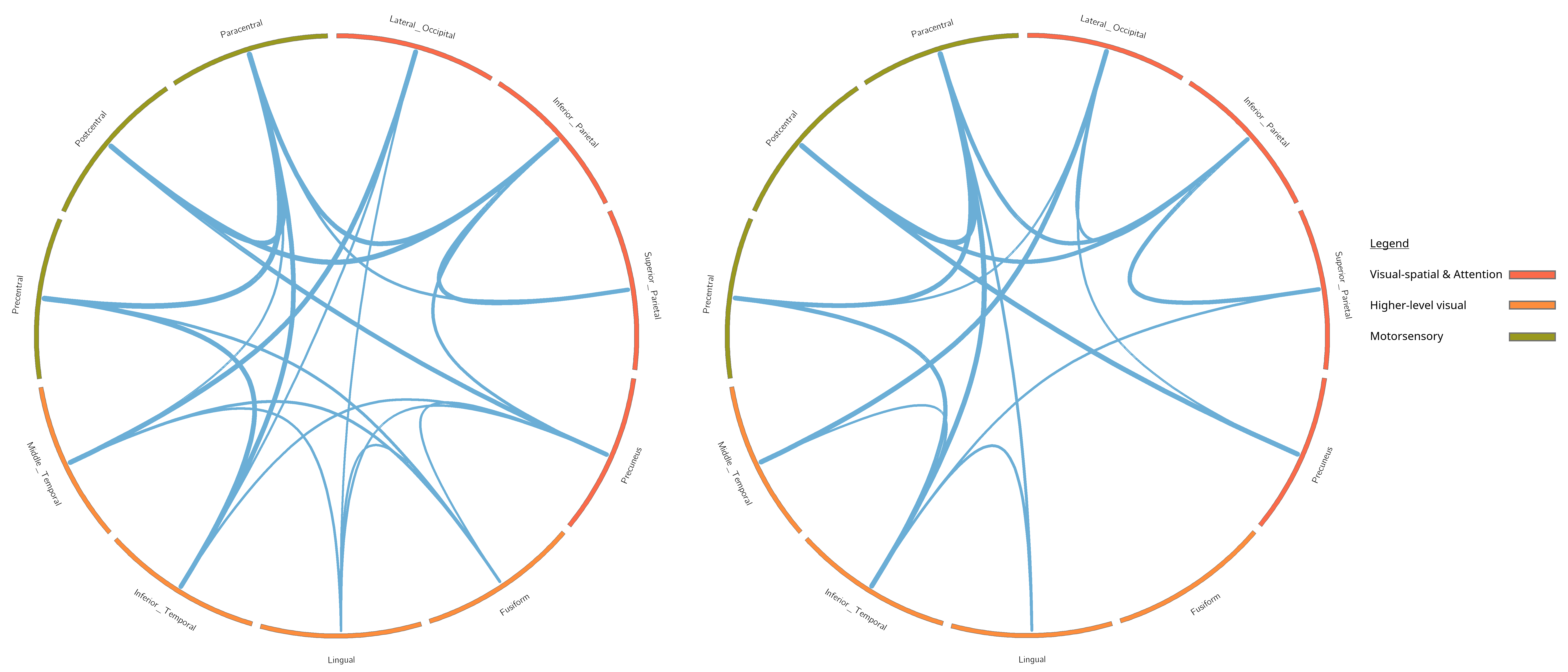}
\caption{
Task activation networks for SINGLE (left) and rt-SINGLE (right) algorithms respectively. Present edges had statistically significant 
positive correlations with task onset after correction for multiple comparisons. 
Edge width is proportional to the magnitude of such correlations.
ROIs are grouped according to their functional description as summarized in 
the legend.
We note there is consistent activation pattern across both algorithms, particularly across nodes 
nodes corresponding to the motorsensory areas.
}
\label{HCPfig}
\end{figure}

Figure [\ref{HCPfig}] shows task activation networks for both the SINGLE and rt-SINGLE algorithms. 
Here edges are only present if they were reported as being significantly correlated with task-evoked HRF function. 
Edge thickness is proportional to the estimated partial correlation between nodes.
We note that there are visible similarities across
each of the algorithms, indicating that the rt-SINGLE algorithm is accurately detecting 
task-modulated changes in functional connectivity.
In particular there are clear similarities in functional connectivity patterns across
the motorsensory regions. 

These results serve as further evidence that the rt-SINGLE algorithm is capable of accurately 
detecting changes in real-time. Formally, the rt-SINGLE algorithm is able to detect patterns 
in functional connectivity associated with a motor task. It therefore follows that 
the estimated functional connectivity networks at each point in time could serve as input when
looking to predict a subject's brain state,
as would be required in BCI applications.

\section{Application: virtual world real-time fMRI data}
\label{sec--Minecraftapp}

While the HCP dataset introduced in Section \ref{sec--HCPapp} serves to demonstrate the reliability of 
the real-time network estimates, 
our proposed method was also tested using data that was processed and studied 
alongside data acquisition. In this proof-of-concept example study, 
we employed a more naturalistic and complex task that is similar to the
type of situation likely to be used in closed-loop BCI systems. 
The study presented in this section involved a subject playing
\textit{Minecraft}, a popular virtual reality game, whilst 
in the scanner. 
The subject was instructed to explore the virtual world, during which time 
the background setting alternated between 
daylight and night. 
Here the objective was to measure if these changes could be reported by the proposed method in real-time. 

\subsection{\textit{Minecraft} game}

\begin{figure}[ht]
\centering
\includegraphics[width=\textwidth]{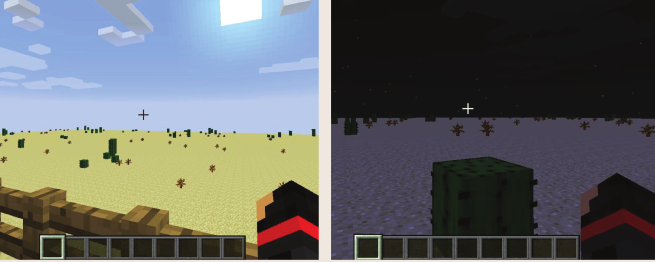}
\caption{Example screenshots of daylight and night from the Minecraft game that subjects are asked to play in the scanner.}
\label{MinecraftPic}
\end{figure}

The subject was instructed to interactively explore a virtual game environment whilst lying in the MRI scanner. 
The game environment was created within the \textit{Minecraft} framework (Mojang AB, Stockholm, Sweden). 
The experiment was repeated five times with 
each run consisting of the subject exploring the virtual world for 
5 minutes (corresponding to 150 TRs). 
During this time, background brightness alternated between daylight and night in a 20 second
blocked fashion. 
An example screenshot of changes seen by the subject is given in 
Figure [\ref{MinecraftPic}].
The subject used button response boxes in both hands to move forward and turn left/right. 
The game was run on a separate machine 
which was connected to the real-time fMRI processing computer via a shared wireless network. 

The study of the dataset is challenging for several reasons;
first, correctly preprocessing and preparing the data in real-time 
is non-trivial and must be implemented efficiently in order to keep up with data acquisition.
Moreover, further challenges 
arise due to the nature of the task performed. 
Formally, the task performed will activate visual, motor as well as higher level cognitive networks.
Estimating such diverse networks with limited observations as well as limited computational resources 
therefore poses a large challenge.

\subsection{Real-time fMRI setup}

Whole brain coverage images were acquired in real-time by a Siemens 
Verio 3T scanner using an EPI sequence
(T2*- of view $192 \times 192 \times 105$ mm, flip angle $80{\degree}$, time repetition (TR) / time echo (TE) = $\nicefrac{2000}{30}$ ms, 35 ascending slices). 
The reconstructed single EPI volume was exported from the MR scanner console to the real-time fMRI processing 
computer (Mac mini, 2.3 GHz Intel Quad Core i7, 16 GB RAM) via a shared network folder. Prior to the online run,
a high-resolution gradient-echo T1-weighted structural anatomical volume (reference anatomical image, RAI with
voxel size $1.00 \times 1.00 \times 1.00$ mm, flip angle $9\degree$, TR / TE = $\nicefrac{2300}{2.98}$ ms, 160 ascending slices, inversion time = 900 ms) 
and one EPI volume (reference functional image, RFI) needed to be acquired. The first step comprised the brain extraction 
of the RAI and RFI using BET (56), followed by an affine co-registration of the RFI to RAI and subsequent linear registration 
(12 DOF) to a standard brain atlas (MNI) using FLIRT \citep{jenkinson2001global}. The resulting transformation matrix was used
to register the 11 anatomical ROIs (as described in Section \ref{data--HCP} and Table \ref{table:MNI}) from MNI to the functional space of the respective subject. 
For online runs, incoming EPI images were converted from dicom to nifti file format and real-time motion correction was 
carried out using MCFLIRT \citep{jenkinson2002improved} with the previously obtained RFI acting as reference. ROI means of the anatomical maps
for each TR were simultaneously extracted using a GLM approach and written into text file that was accessed by the 
rt-SINGLE algorithm. A burn-in period of 10 observations was employed.
The first of the five runs was used to estimate sparsity and temporal homogeneity parameters by minimizing AIC. These 
parameters were subsequently used in the remaining four runs. Adaptive filtering was employed to estimate subject 
covariance matrices with tuning parameter $\eta=0.005$.
Preprocessing together with the rt-SINGLE optimization required under one second of computational time,
making the proposed algorithm feasible within a neurofeedback setting.

\subsection{Results}
For each TR, updated ROI time courses were 
studied in real-time using the rt-SINGLE algorithm. 
Adaptive forgetting was employed with $\eta=0.005$ and regularization parameters where estimated by minimizing AIC over the subjects first run, these values where 
subsequently fixed throughout the remaining four runs.

\begin{figure}[ht]
\centering
\includegraphics[width=\textwidth]{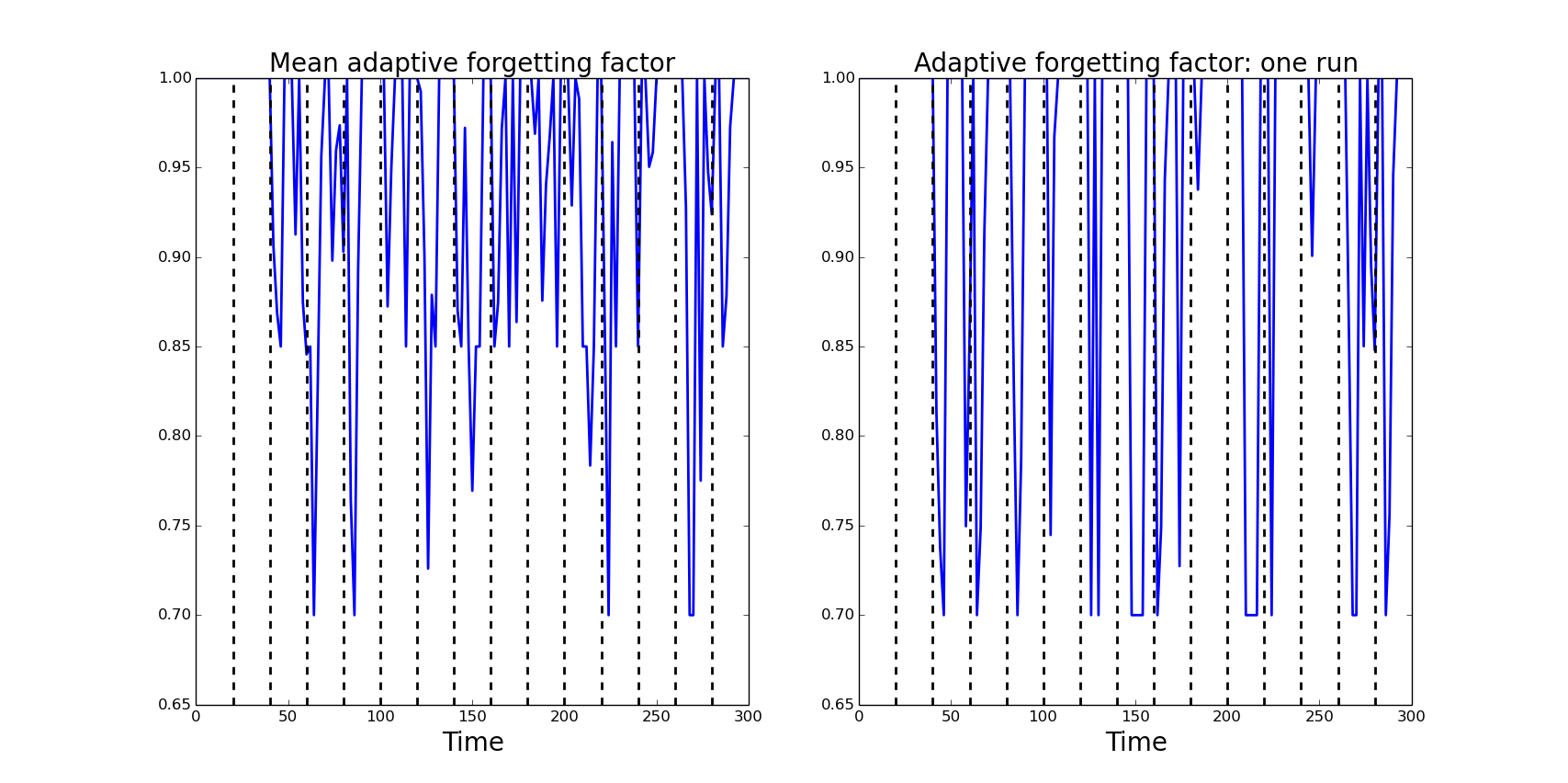}
\caption{ Mean adaptive forgetting factor, $r_t$, over all four runs (left) and over a single run (right). 
The vertical dashed lines indicate times when the background changed from daylight to night or vice versa. 
We note that there is a recurrence of the forgetting factor dropping slightly after changes occur. }
\label{adaptiveFig}
\end{figure}

As discussed previously, one of the additional benefits of adaptive forgetting is that 
further information can be gathered by studying 
the value of adaptive forgetting factor, $r_t$, over time. 
Large values of $r_t$ are indicative of piece-wise stationary connectivity structures while small values serve to denote a period of instability. More importantly, sudden
drops in the value of $r_t$, as shown in Figures [\ref{simSF}] and [\ref{simSW}], can serve as a suggestion
that a change has occurred and that the algorithm is quickly adapting. 
Figure [\ref{adaptiveFig}] shows both the mean adaptive forgetting factors over all four
runs as well as the adaptive forgetting factor
estimated for a single run. The vertical dashed lines indicate when the background was changed from daylight to night or vice versa. 
We note that the forgetting factor quickly drops after most of these changes as we would expect. 
Moreover, there are no drops present during the first 20 seconds as this corresponds to the burn-in period employed.

\begin{figure}[ht]
\centering
\includegraphics[width=.8\textwidth]{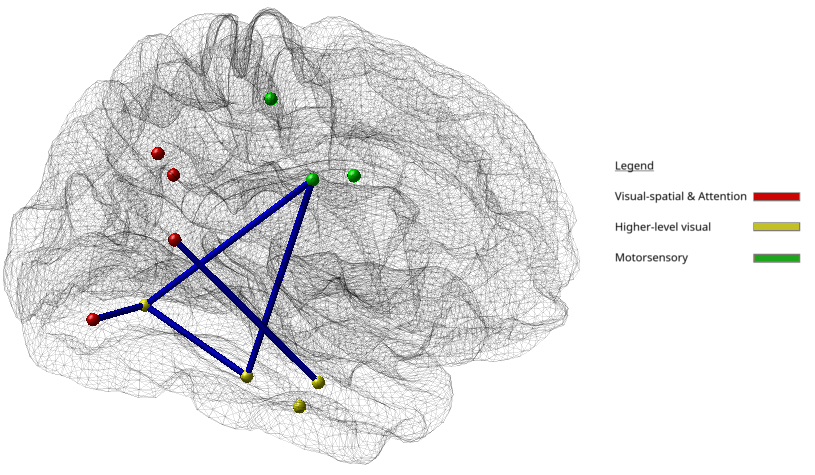}
\caption{ Visualization of daylight modulation network. Present edges are activated where significantly activated during 
the daylight blocks. There is a network for the activated edges involving both the 
Lingual gyrus as well as the Fusiform gyrus regions
which are typically associated with higher-order visual processing.
}
\label{dayNightActivation}
\end{figure}

The estimated networks can be used to study how 
functional connectivity is affected by changes related to daylight. By quantifying differences in the estimated 
functional connectivity networks we are able to report the edges that are indicative of daylight and night respectively. 
Figure [\ref{dayNightActivation}] shows the edges that were significantly activated during daylight blocks. 
To determine the statistical significance of the reported changes a Wilcoxon rank-sum test was employed
and the resulting $p$-values where adjusted to correct for multiple comparisons using 
the Holm-Bonferroni method \citep{holm1979simple}.

Figure [\ref{dayNightActivation}] plots the statistically significant changes in
functional connectivity modulated by changes in daylight settings. 
In particular, an
increase in connectivity was detected during daylight blocks
for visual-spatial and higher-level visual areas as well as higher-level visual and motorsensory areas.
This is evident in Figure [\ref{dayNightActivation}] which 
shows a clear hub of activated functional edges
centered around the Lingual gyrus and the Fusiform gyrus regions;
two regions 
typically associated with higher-order visual processing.
It follows that information of this nature 
could be used to dictate feedback to subjects in real-time or as part of a 
BCI interface in future. Moreover, more advanced
machine learning classifiers could be employed as described in \cite{laconte2007real}.

\section{Discussion}
\label{sec--Discussion}

In this work we introduce a novel methodology with which to estimate dynamic functional connectivity networks in
real-time. The strengths of the proposed method can be summarized as follows.
First, the proposed method may leverage adaptive forgetting methods in order to obtain highly adaptive estimates of 
the sample covariance over time. Such methods designate that choice of the forgetting factor to the data, making them
highly adaptive as well as flexible. 
The latter point is of particular importance in the rt-fMRI setting; 
since changes in functional connectivity may
occur abruptly and at varying intervals
the assumptions behind a fixed forgetting factor do not necessarily hold true. 
Second, by extending the recently proposed SINGLE algorithm we are able to accurately estimate functional connectivity networks based on precision
matrices in real-time. 
The proposed method enforces constraints on both the sparsity as well as the temporal 
homogeneity of estimated functional connectivity networks.
The former is required in order to ensure the estimation problem remains
well-posed when the number of relevant observations drops, as is bound to 
occur when adaptive forgetting is employed.
On the other hand, the temporal homogeneity
constraint ensures changes in functional connectivity are only reported when heavily substantiated by evidence in
the data. 
As a result, the rt-SINGLE algorithm is able to both obtain accurate estimates of functional
connectivity networks at each point in time as well as 
accurately describe the evolution of networks over time.

The rt-SINGLE algorithm is closely related to sliding window methods which have
been employed extensively in the real-time setting 
\citep{gembris2000functional, esposito2003real, ruiz2014real, zilverstand2014windowed}.
Extensions of sliding window methods, such as EWMA models, have been successfully 
applied to offline fMRI studies \citep{lindquist2007modeling}
and have been shown to be better suited to estimating dynamic functional connectivity \citep{lindquist2014evaluating}. 
In this work EWMA models are considered alongside adaptive forgetting.
The latter can be interpreted as a natural extension of EWMA models, where the rate at which past observations 
are discarded is learnt from the data.
The proposed method is flexible and can be implemented using either a 
fixed forgetting factor (corresponding to an EWMA model) or using adaptive forgetting. 

The proposed method requires the input of three parameters. 
The first of these parameters affects the manner in which sample covariance matrices are estimated. 
As noted previously, this can be achieved either using an EWMA model or via adaptive forgetting.
If the former is used a fixed forgetting factor, $r$, must be specified. 
As discussed in Section \ref{sec--rtSINGLEParams},
the choice of $r$ can be interpreted as defining a weighted window.
It therefore follows that the choice of $r$ must balance a 
trade-off between stability and  adaptivity. A small choice of $r$ implies past
observations are quickly discarded. While this
will result in highly adaptive estimates, it may also result in network estimates that are dominated by noise. Conversely,
a large value of $r$ will diminish the adaptive properties of the proposed method
whilst producing more stable estimates. 
Alternatively, the use of adaptive forgetting requires the input of a stepsize parameter $\eta$. This parameter
governs the rate at which an adaptive forgetting factor, $r_t$, varies and can be interpreted as the
stepsize in a stochastic gradient descent scheme \citep{bottou2004stochastic}. As such, 
it is typically suggested to set $\eta$ in the range of $0.001$ to $0.05$.
The final two parameters enforce sparsity and temporal homogeneity respectively. 
These parameters remain fixed throughout in a similar manner to the fixed forgetting factor
and two heuristic approaches are proposed to tune these parameters. 
A future improvement for the proposed algorithm would involve adaptive regularization
penalties. However, such approaches are computationally and theoretically challenging due
to the non-differentiable nature of the penalty terms.

Several simulations are presented to examine the properties of the proposed method.
These serve to demonstrate that the proposed method is capable 
of accurately estimating functional connectivity networks in real-time. 
Formally, we investigate three specific properties of the proposed method. First, 
in simulation \rom{1}, we quantify
how effectively the proposed method can 
track changes in
covariance structure over time. Second, simulations \rom{2} and \rom{3} we study
how accurately the rt-SINGLE algorithm
can estimate functional connectivity networks. 
Finally, the computational cost of the proposed method is considered in simulation \rom{4}. This is 
fundamental in the case of real-time algorithms as estimated networks must be reported for each observation. 

Two applications of the proposed method are provided. The first involves motor-task data from the HCP. 
While this data is not intrinsically real-time, it is included as a proof-of-concept 
study to validate the proposed method. The results demonstrate that the rt-SINGLE algorithm is able to 
accurately detect functional networks which are modulated by motor task. 
The second application corresponds to a more complex real-time study which is more closely related to the 
type of situation which may be employed in a closed-loop BCI system.
This study required the subject to explore a virtual game environment while in the scanner.
Throughout this time, the background 
brightness alternated between daylight and night in a block fashion. The 
changes in background brightness induced changes in functional connectivity 
that were subsequently reported, in real-time, by the proposed method.
Specifically, the proposed method is able to detect a network of edges that was activated during daylight. 
In future, such information could be incorporated into a neurofeedback or BCI setting. 

In conclusion, the rt-SINGLE algorithm provides a novel method for estimating 
functional connectivity networks in real-time. 
We present two applications demonstrating that the rt-SINGLE algorithm is capable of reporting 
changes in functional connectivity related to changes in task as well as external stimuli. 
In future, we look forward to incorporating the proposed method in a more extensive real-time 
neurofeedback study.

\newpage
\section*{Appendix}
\renewcommand{\thesubsection}{\Alph{subsection}}
\label{app2}

\subsection{Full derivation of Adaptive filtering derivative}
The results shown in this section are taken from \cite{anagnostopoulos2012online}.

The log-likelihood for unseen observation, $X_{t+1}$ is given by 
\begin{equation}
 \mathcal{L}_{t+1} = \mathcal{L} (X_{t+1}; \bar x_t, S_t) = -\frac{1}{2} \mbox{ log det} (S_t) - \frac{1}{2}(X_{t+1} - \bar x_t)^T S^{-1}_t (X_{t+1} - \bar x_t).
\end{equation}
The approach taken here is to approximate the derivative of $\mathcal{L}_{t+1}$ with respect to adaptive forgetting factor $r_t$
by calculating the exact derivative of $\mathcal{L}_{t+1}$ with respect to a 
fixed forgetting factor $r$. Then under the assumption that changes in $r_t$ occur sufficiently slowly, this will serve
as a good approximation to the derivative of $\mathcal{L}_{t+1}$ with respect to  $r_t$.

We begin by noting the following results from \cite{petersen2008matrix}:
\begin{align}
\label{eq1}
 \frac{\partial ~\mbox{log det}~ (S_t)}{ \partial r} &= \mbox{Trace}~(S_t^{-1} S_t')\\
 \frac{\partial  (S_t^{-1})}{ \partial r} &=  - S_t^{-1} S_t' S_t^{-1}.
\end{align}
Moreover, we note that we do not need to explicitly invert $S_t$. By noting that $S_t$ is a rank one update
of $S_{t-1}$ we are able to directly obtain $S_t^{-1}$ using the Sherman-Woodbury formula.

Further, from equations (\ref{EWMAmu}), (\ref{EWMApi}), (\ref{EWMAcov}) and (\ref{EWMAomegaAdaptive}) we can see that:
\begin{align}
 \bar{x}_t' &= \left (1 - \frac{1}{\omega_t} \right ) \bar{x}_{t-1}' + \frac{\omega_t'}{\omega_t^2} \left (  X_t - \bar{x}_{t-1} \right )\\
 \omega_t' &= r_{t-1} \omega_{t-1}' + \omega_t \\
 \Pi'_t &= \left (1 - \frac{1}{\omega_t} \right ) \Pi_{t-1}' + \frac{\omega_t'}{\omega_t^2} \left (  X_t X_t^T - \Pi_{t-1} \right ) \\
 \label{eqlast}
 S_t' &= \Pi_t' - \bar{x}_t' \bar{x}_t^T - \bar{x}_t (\bar{x}_t')^T,
\end{align}
where once again we have used the notation $A'$ to denote the derivative of a vector or matrix A with 
respect to $r$.
Using the results from equations (\ref{eq1}) to (\ref{eqlast}) we can directly differentiate the $\mathcal{L}_{t+1}$ to obtain
equation (\ref{likelihoodDerivative}).

\newpage
\bibliographystyle{plainnat}
\bibliography{ref}
\end{document}